\documentclass{article}

 \usepackage[preprint]{neurips_2025}

\usepackage[utf8]{inputenc} %
\usepackage[T1]{fontenc}    %
\usepackage{url}            %
\usepackage{booktabs}       %
\usepackage{amsfonts}       %
\usepackage{nicefrac}       %
\usepackage{microtype}      %
\usepackage{xcolor}         %
\usepackage{graphicx}
\usepackage{wrapfig}
\usepackage{enumitem} %

\usepackage{graphicx}
\usepackage{booktabs}
\usepackage{comment}

\usepackage{arydshln} %

    \usepackage{pifont}

\usepackage[accsupp]{axessibility}  %

\usepackage{listings}
\usepackage{xcolor}
\usepackage{caption}
\usepackage[pagebackref]{hyperref}
\usepackage[capitalize]{cleveref}
\usepackage{subcaption}

\title{ProcFunc: Function-Oriented Abstractions for Procedural 3D Generation in Python}

\author{%
\textbf{
Alexander Raistrick \quad
Karhan Kayan \quad
Jack Nugent \quad
David Yan} \\
\textbf{
Lingjie Mei \quad
Meenal Parakh \quad
Hongyu Wen \quad
Dylan Li} \\
\textbf{
Yiming Zuo \quad
Erich Liang \quad 
Jia Deng} \\
Princeton University, Department of Computer Science \\
\texttt{\{araistrick, karhan, jacknugent, yan.david, lm5483,} \\
\texttt{meenalp, hongyu.wen, dl2223, zuoym, erliang, jiadeng\}@princeton.edu}
}

\begin{document}

\maketitle

\begin{abstract}

We introduce ProcFunc, a library for Blender-based procedural 3D generation in Python. ProcFunc provides a library of easy-to-use Python functions, which streamline creating, combining, analyzing, and executing procedural generation code. ProcFunc makes it easy to create large-scale diverse training data, by combinatorial composition of semantic components. VLMs can use ProcFunc to edit procedural material and geometry code and can create new procedural code with significantly fewer coding errors. Finally, as an example use case, we use ProcFunc to develop a new procedural generator of indoor rooms, which includes a collection of new compositional procedural materials. We demonstrate the detail, runtime efficiency, and diversity of this room generator, as well as its use for 3D synthetic data generation. Please visit \href{https://github.com/princeton-vl/procfunc}{this https url} for source code.
\end{abstract}

\section{Introduction}
\label{sec:intro}

Procedural 3D generation refers to the creation of 3D assets---objects, scenes, materials, textures, animations---through algorithms and compact mathematical rules, in other words, human-understandable code. For example, the 3D mesh of a tree can be created by randomized recursive branching starting from a base cylinder; the texture map of the tree bark can be created as a simple function of a grid of random 2D vectors~\citep{perlin1985image}. 

The primary advantages of procedural generation are control and composition. In essence, procedural generation represents 3D assets as code; it thus allows precise and fine-grained control, from the curvature of a tree leaf to the layout of a forest. Because procedural generators are code, they can be modular and composable. By composing modules and varying control parameters, procedural generators can create unlimited, diverse, varied, and photorealistic 3D assets, as demonstrated by Infinigen~\citep{infinigen2023infinite,infinigen2024indoors}, a Blender-based procedural system. Procedural generation has proven useful in various ways, including creating synthetic training data \citep{yan2026makesgoodsynthetictraining}, evaluation \citep{evalrobustdepth,infinibench}, and AI-assisted content creation \citep{huang2024blenderalchemy,llm3d}. 

\begin{figure}[tb]
  \centering
  \includegraphics[width=\textwidth]{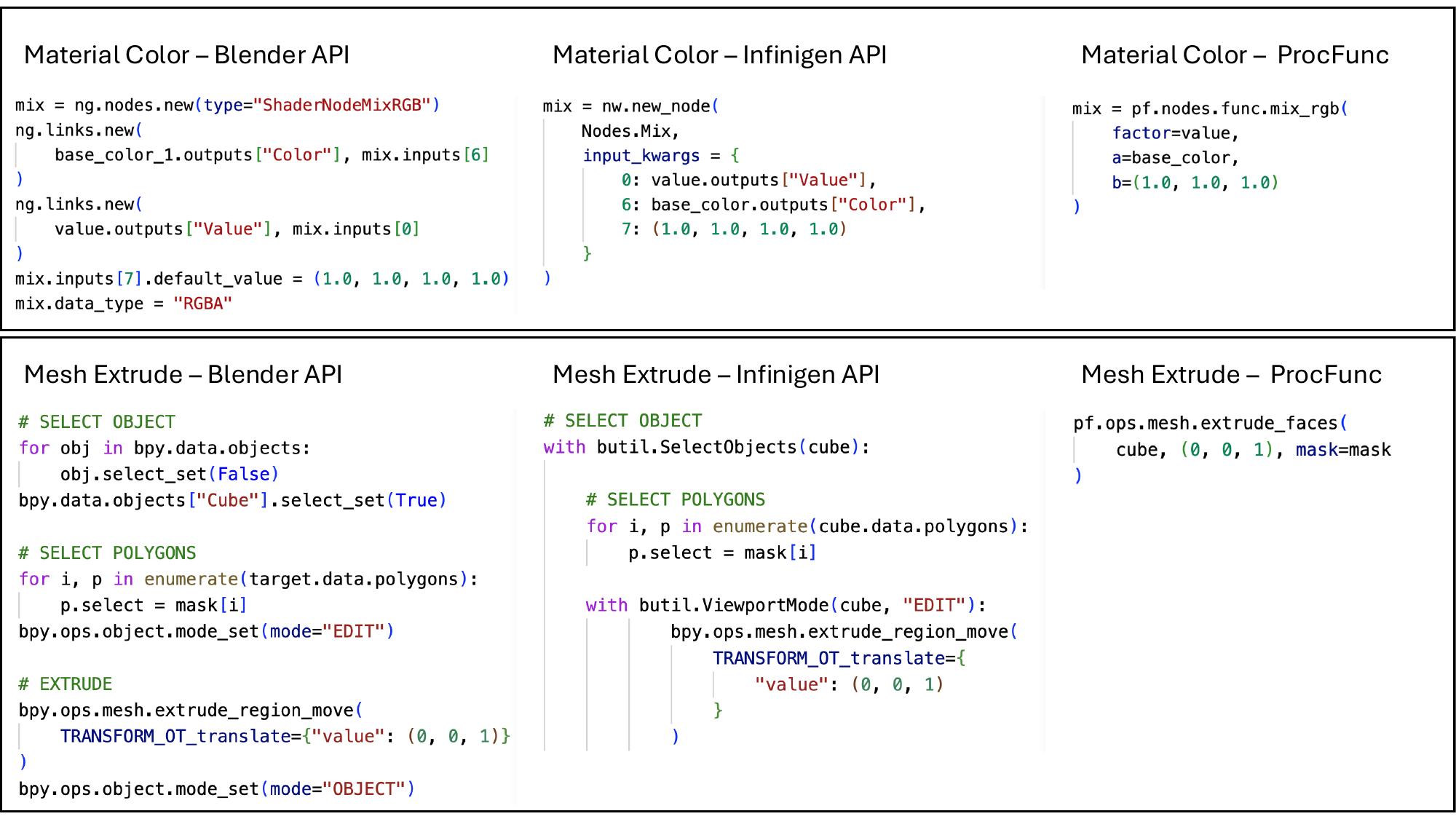}
  \caption{\textbf{ProcFunc Primitives API} - ProcFunc provides a new API for low-level graphics operations which eliminates all non-essential or GUI-oriented complexity. Typical usage of the Blender and Infinigen API requires separate selection and action steps to manipulate Blender's global background state. Blender and Infinigen require unclear runtime keys such as ``0, 6, 7'' in the top row, which change dynamically depending on other arguments or program state. ProcFunc instead provides each operation as an atomic Python function with all inputs as explicit arguments. }
  \label{fig:code_example}
  \vspace{-1em}
\end{figure}

Because procedural generators are code, the development of procedural generators can benefit greatly from well-designed and easy-to-use programming abstractions, just as deep learning has benefited greatly from libraries such as TensorFlow, PyTorch, and JAX. A good programming abstraction can reduce implementation complexity, improve development efficiency, and enable new applications.

A good programming abstraction for procedural generation should allow efficient creation and composition of procedural generators in the form of compute graphs that define shape or texture, similar to how PyTorch can define and combine neural modules through a few lines of code. Also similar to PyTorch, it would be highly useful to be able to easily access, modify, analyze the control parameters to enable precise control and domain randomization, which would be useful in both content creation and training data generation.   

However, existing open-source programming interfaces for procedural generation fall short of these desiderata. For example, composing mathematical functions is a basic operation in procedural generation, but often requires lengthy code to implement using Blender's native Python API. It takes multiple steps to create a gadget and connect its input and output sockets, and particular sockets may appear or disappear depending on the gadget state and may have dynamic indices and duplicated names. 
To be clear, this does not mean that the Blender API is flawed, just that it is designed for a different purpose, namely to facilitate a GUI-based workflow: these steps are intuitive for artists and GUI-centric plugins, but can be significantly simplified for Python-centric development.

Infinigen's code streamlines some of these steps, but does not fundamentally reduce their complexity. Each operation still requires multiple steps (albeit wrapped by new functions), and still depends on dynamically changing UI strings Fig.~\ref{fig:code_example}. This leads to complex procedural code, where one function can fail silently due to the global state configured by a previous function, and modules are fragile and hard to combine. In addition, the control parameters lack standardization and are hard to access, making it difficult to perform arbitrary customization and randomization. 

In this work, we introduce ProcFunc, a library for efficient creation, composition, analysis and execution of Blender-based procedural generators. ProcFunc automates common tasks in creating procedural generators, and makes generators easy to combine and re-parameterize. Procedural generators using ProcFunc support many research tasks without complex per-asset implementation: we provide tools which can automatically analyze generators to access their parameters and composition, access intermediate values, and analyze or optimize their overall random distribution. We aim to streamline the overall process of procedural dataset generation, much in the same way that deep learning libraries \citep{tensorflow,pytorch,jax2018github} have streamlined the process of creating and using deep learning models.

\begin{figure}[tb]
  \centering
  \includegraphics[width=\textwidth]{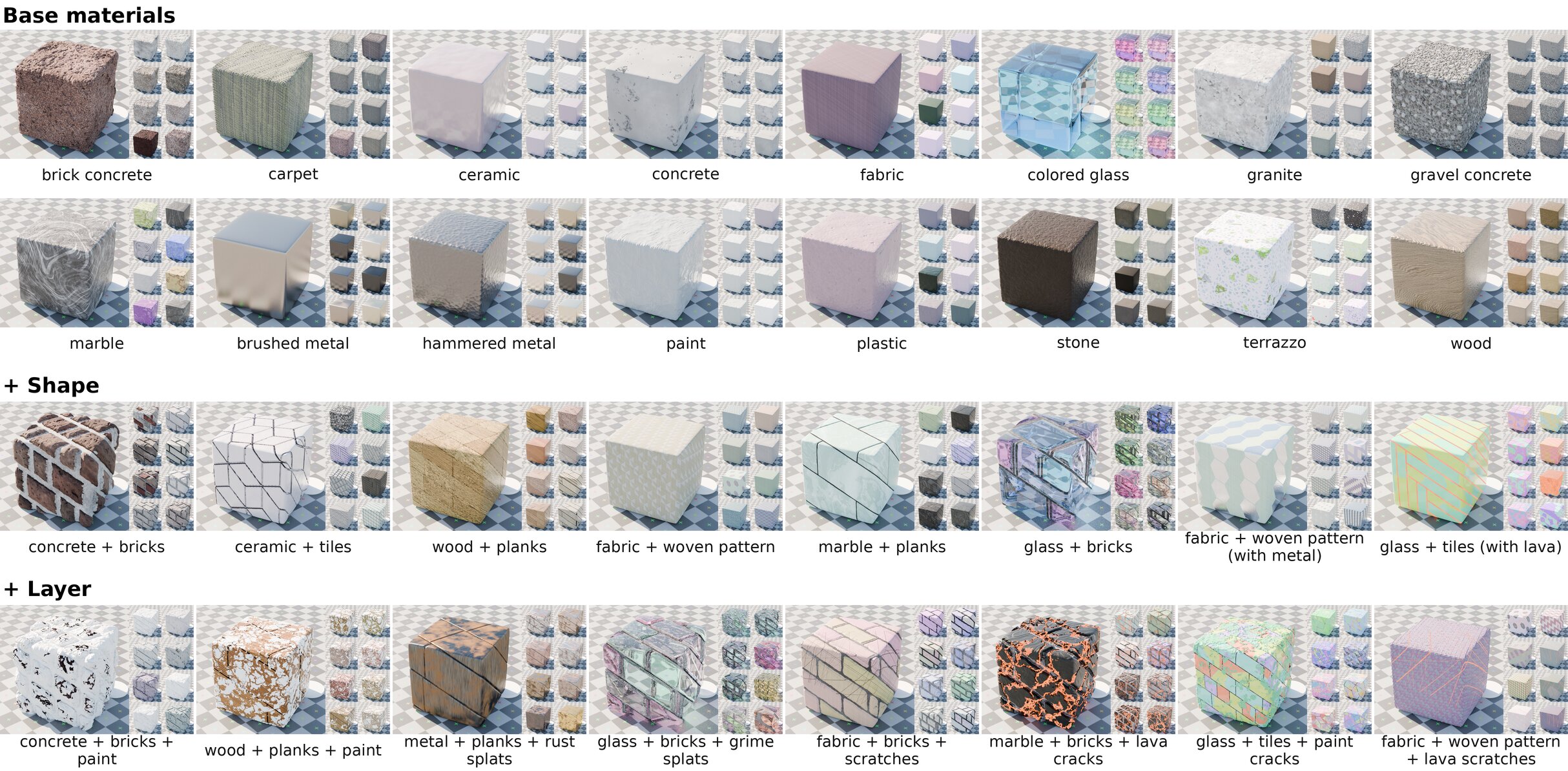}
  \caption{\textbf{Compositional Procedural Generation} -  ProcFunc enables reusable procedural generators which can form countless semantically meaningful combinations. For materials, one can create bricks, tiles, paint and scratches combined with any material, forming tens of thousands of combinations.}
  \label{fig:materials}
  \vspace{-1em}
\end{figure}

ProcFunc is particularly useful in increasing \textit{compositionality}. Compositionality is the degree to which a system can be decomposed into a hierarchy of reusable parts, and recombined (composed) into valid and useful combinations. This is helpful for procedural generation because recombining components can drastically increase visual diversity and coverage (Fig.~\ref{fig:materials}). 

Compositionality and ease of use depend on newly designed APIs at three levels. First is simple, explicit functions for all graphics operations in Blender, which significantly simplify Python-oriented procedural generation, and can be used for individual asset generation or content creation. Second is a design pattern for module-level composition of procedural generators, which provides individual assets as off-the-shelf functions with either precise control or automatic randomization, and in some cases can be statically analyzed using a compute graph tracer. Finally, building on top of ProcFunc, we provide a library of high-quality procedural generators. It features compositional materials (Fig.~\ref{fig:materials}), as well as simple tools for object arrangement and collisions which can be used to efficiently create full rooms (Fig.~\ref{fig:livingrooms}) or user-designed scenes.

We evaluate ProcFunc's effectiveness for its two primary applications: instance level procedural generation for content creation, and useful procedural data \textit{distributions} to be sampled for synthetic data. For content creation, we use BlenderGym \citep{blendergym}'s material and geometry tasks. ProcFunc is helpful in both the default parameter-editing setting and a new from-scratch generation setting, albeit with a non-standard harness. For synthetic dataset generation, we demonstrate our library of components via a new diverse and high-quality compositional indoor room generator. We find it provides greater analytical diversity and semantic coverage due to composition, offers a favorable performance-complexity tradeoff, and is useful in creating large synthetic datasets.

In summary, we provide ProcFunc, a system which substantially reduces the complexity of creating and combining procedural generators. Using it, we create an indoor room generator which has diverse, detailed and compositional materials, and is efficient and useful for dataset generation.

\section{Related Work}

\paragraph{Procedural Dataset Generators} Computer vision has benefited from procedural generation at different levels, including scene-level procedural arrangement of fixed objects \citep{procthor}, or procedural variation within objects \citep{scenesynthesizer,artipg}, or fully procedural scenes \citep{infinigen2023infinite,infinigen2024indoors}. Our work complements these by providing new tools and new compositional components which can be combined with any existing system. It is also distinguished by providing a new programming interface, and by extreme compositionality (6 or more layers of modularity within materials, as opposed to at most 2 in Infinigen).

\paragraph{Automated Graphics Editing} Many works combine LLMs and VLMs with Blender or other graphics systems. This is often for scene arrangement of non-procedural objects and materials \citep{fireplace,holodeck,sun2025layoutvlmdifferentiableoptimization3d,yang2025sceneweaverallinone3dscene,layoutgpt}, but has also been applied for high-level configuration of existing procedural generators \citep{3dgpt,worldcraft,code2world,zhou2025roomcraftcontrollablecomplete3d}. Most directly related to our work is automated graphics editing at the object and material level \citep{huang2024blenderalchemy,llm3d,li2025vlmaterialproceduralmaterialgeneration}. However, ProcFunc's primitives are a complementary contribution to all LLM/VLM systems: we provide a new graphics API and high quality procedural generator examples, which can improve any LLM/VLM system, rather than advancing agent or LLM design.

\paragraph{Domain-specific Languages for Procedural Generation} Languages for scene generation and procedural content creation vary in scope. The Scene Language \citep{zhang2025scenelanguage} and Procedural Scene Programs \citep{openuniverse} create useful scene-level arrangements, including control flow and symmetry. However, these rely on pre-made non-procedural object and material assets, whereas ProcFunc's APIs can be composed to create entire scenes using similar tools as would be used by an artist. Other works propose languages \citep{ganeshan2024parselparameterizedshapeediting,jones2026shapelibdesigninglibraryprogrammatic} or generative grammars \citep{10.1145/1141911.1141931} for shapes at the object level, especially for object editing or abstract shape generation. ProcFunc has a broader scope of primitives, by including every procedural Blender operation (materials, extrusion, vertex positioning, bezier surfaces, etc.), and can tackle from-scratch procedural generation. These graphics operations can implement scenes more compactly than large numbers of shape primitives.

Most similar to our tools are the operator and geometry node tools introduced in Infinigen \citep{infinigen2023infinite}, which are also used for material and geometry editing tasks in BlenderGym \citep{blendergym}. Compared to these tools, ours requires less knowledge of the Blender UI (Fig.~\ref{fig:code_example}), as it is fully specified as static Python functions, and is more concise and less error-prone (Tab.~\ref{tab:fromscratch}).

\section{ProcFunc}

ProcFunc's overall goal is to make procedural generators easy to create and use in Python, especially for content creation and synthetic dataset generation. This includes creating procedural generators from scratch, quickly combining existing generators, customizing dataset distributions \citep{wen2025layereddepth} and meeting task requirements such as paired synthetic data \citep{evalrobustdepth} or providing full parameter access \citep{inverting}. 

\subsection{Primitives API}

We maximize the simplicity of combining primitive graphics operations while ensuring they are capable of generating any procedural scene. This results in a carefully designed set of 497 Python functions which express the same functionality as commonly used procedural tools in Blender, such as material and mesh editing operations.

We design each primitive to expose a single graphics algorithm as an atomic function, even if the corresponding Blender construct is multiple functions, or is many algorithms combined into a single function. Instead of a single ``Math'' shader function we provide many arithmetic functions, or for a single ``Noise'' function we provide a separate function with different arguments for each type of noise. We also combine functions, so that users do not need different functions for different data types or for materials vs geometry. We ensure each function exposes all dependencies as explicit function arguments, rather than implicit background state, and returns all its effects as return values, including any active faces as NumPy arrays. ProcFunc's function and argument names are similar to Blender's operation names, but are changed for clarity and disambiguation (e.g. duplicate ``Vector'' arguments in Fig.~\ref{fig:code_example}).

Each primitive function is executed by configuring the correct context and executing the operation through the Blender Python API. We use the most efficient possible invocation of the underlying operations. The exact implementation of each primitive can also evolve in future versions of ProcFunc without affecting its interface. At execution-time we also perform runtime type inference of ``data\_type'' arguments (see Fig.~\ref{fig:code_example}), and resolve operators ``\texttt{+-/*//}'' to shader or other arithmetic with that datatype. 

Finally, we provide a \textit{transpiler}, which converts procedural generators from Blender Nodegraphs to Python code to be used with LLMs or other code. Our transpiler is similar to Infinigen's, but adapted to ProcFunc's interface. It also inlines short expressions, uses operator overloads, and can post-process graphs to use HSV colors and interpolate example parameters. This allows Python code to assist in GUI-based editing, and is essential for LLMs to edit pre-existing Blender nodegraphs in ProcFunc. See Appendix~\ref{sec:supp_impl} for example ProcFunc material programs.

\subsection{Module Composition Framework} 

All asset generators in ProcFunc are simple functions which are easily controllable by setting their input arguments. These functions generate and return a useful asset with no other steps or configuration. We define standardized function interfaces for categories of generators (\textit{Masks, Materials, Objects, Scenes}), which guarantee certain inputs / output args will be present, such as Materials having a ``vector'' coordinate system input, and return a ``surface'', ``displacement'' and ``volume'' to define shading and texture. This enables widespread compositionality, as generators can be substituted while providing the same interface. Our primitives API is also essential to this, as otherwise there are implicit inputs and outputs which are challenging to capture in an interface. 
\paragraph{Random Generation} We also distinguish between random generators and deterministic generators. Each class of generator (e.g. Wood) has one or more \textit{random sampler functions}, which sample parameters then create and return the asset. Random samplers are useful for dataset generation or any case in which the user doesn't wish to specify exact parameters. There can be multiple samplers for different distributions of a single deterministic asset. All random functions require a ``np.random.Generator'' (RNG) input, so randomness is explicitly declared and reproducible from seed, and also ensures no deterministic generator can call a random one.

\paragraph{Static Analysis with Program Tracing} Our design, so far, is simple but not accessible for inspection or optimization, as random parameters are local to the functions that use them. We address this with a tracer, similar to JAX and especially PyTorch's \texttt{fx.symbolic\_trace}, which executes a program in a special context to capture its compute graph without executing any expensive operations. Our tracer can produce both \textit{Instance-level Graphs} and \textit{Distribution-level Graphs}. Instance-level Graphs capture only the control-flow path and numeric parameters that affected a single output. \textit{Distribution-level} graphs capture random distributions (uniform, normal, etc) of parameters, and capture all possible discrete combinations in the graph, to allow dataset-level optimization of random choices. 

Both types of graphs can also be converted into Python code corresponding to a particular asset, with all parameters and public function calls specified - see Appendix~\ref{sec:supp_tracer} for examples. Overall, each random procedural sampler can produce unlimited specific instance-level graphs with concrete parameters corresponding to an image. We anticipate these will be useful for precise per-asset control or as training supervision for image-to-program / inverse procedural generation tasks.

\subsection{Pre-implemented Procedural Generators}
\label{sec:basegen}

\begin{figure}[tb]
  \centering
  \includegraphics[width=\textwidth]{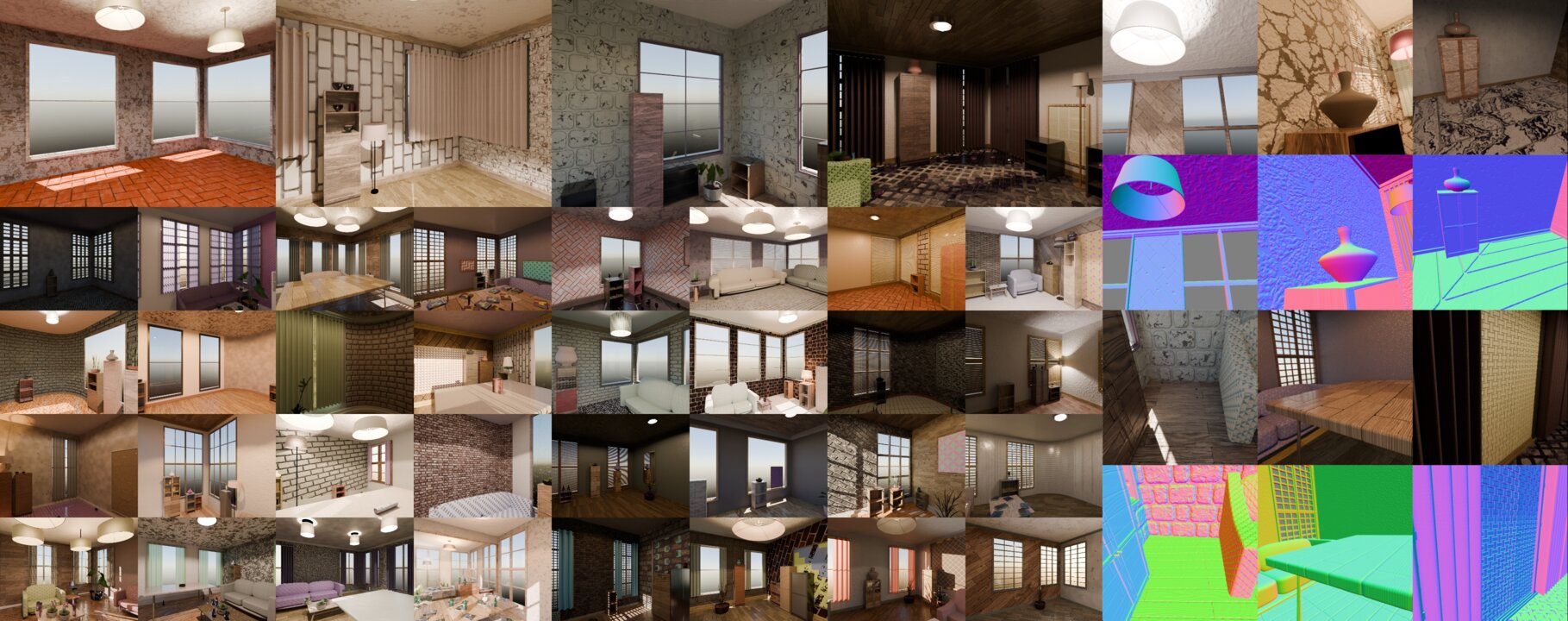}

  \caption{\textbf{Example Room Generator} - We use ProcFunc materials and scene generation tools to create procedural indoor rooms.}
  \label{fig:livingrooms}
  \vspace{-1em}
\end{figure}

We provide a high-quality library of pre-made procedural generators implemented using ProcFunc, as well as tools to easily customize and make datasets using these generators. We use these to create an off-the-shelf indoor room generator, shown in Fig.~\ref{fig:livingrooms}.

We provide 35 newly-created high-quality procedural material components:
\label{sec:premade}
\begin{itemize}[itemsep=0.1pt, topsep=0pt,leftmargin=*]
\item \textit{Material Base} - Concrete, Ceramic, Metal, Fabric, Glass, Granite, Gravel, Marble, Paint, Plastic, Stone, Terrazzo and Wood.
\item \textit{Material Shapes} - Tiles (10 variants), Bricks (4 variants), Masonry, Planks
\item \textit{Material Masks} - Cracks, Flakes, Smudges, Dirt, Scratches, Edge Wear
\end{itemize}

Random samples and compositions are shown in Fig.~\ref{fig:materials} and Appendix~\ref{sec:supp_materials}. Each category has significant internal diversity. ProcFunc's interfaces enable arbitrary compositions of these modules, with thousands or tens of thousands of useful semantic combinations depending on the desired granularity of categories. We also provide altered versions of 14 object-level procedural generators adapted from Infinigen Indoors, which are refactored to meet our interface by accepting procedural parameters and externally provided materials as function arguments.

\paragraph{Scene and Dataset-Generation Tools} We provide several easy-to-use tools which enable simple Python scripts to create new scene arrangements and dataset generation pipelines:
\begin{itemize}[itemsep=0.1pt, topsep=0pt,leftmargin=*]
    \item \textit{Collision Checking} - We use Python-FCL \citep{pan2012fcl} but efficiently re-use bounding-volume-hierarchies for any duplicate objects or rigid motion between queries.
    \item \textit{Iterative Snapping} - we align objects front-to-back or side-to-side with simple function calls, and provide a framework for concise rejection sampling. This enables easy creation of new scene types.
    \item \textit{UV Surface Cutouts} - we create walls by aligning UV-surface handles to windows/door edges instead of via mesh boolean. This preserves quadrilateral mesh topology, which then allows efficient subdivision to create dense meshes, rather than requiring volumetric marching cubes. 
    \item \textit{Camera Trajectories} - we implement efficient \textit{Rapidly-exploring Random Tree} (RRT*) \citep{karaman_sampling-based_2011,LaValle1998RapidlyexploringRT} trajectories, same as in TartanAir \citep{wang2020tartanairdatasetpushlimits}. 
    \item \textit{Rendering and Ground Truth} - we provide simple functions for Blender Cycles \& EEVEE rendering and ground truth. We support any annotation type and NumPy output format. We also optionally compress ground-truth storage cost by 75\% via bit-casting and video compression  \citep{ffmpeg}.
\end{itemize}

We use these tools to create a procedural indoor room generator (Fig.~\ref{fig:livingrooms}) and augment it with floating objects and lights for stereo datasets (Sec.~\ref{sec:stereo}). See Appendix~\ref{sec:supp_basegen} for discussion of each tool.

\section{Experiments}

We evaluate ProcFunc's ease of use for creating and editing procedural generators. We use VLM editing due to the availability of standardized quantitative benchmarks. We expect the results to correlate to usefulness of content creation or LLM-assisted coding, and serve as a proxy for human-understandability and similarity to the general distribution of Python code. 
We also demonstrate ProcFunc's usefulness for creating new procedural dataset generators. Particularly, we investigate their semantic coverage, analytical diversity, runtime efficiency, level of ground truth detail, and use for model training.

\subsection{Procedural Parameter Editing}

We compare the relative success of VLM models when using ProcFunc to solve BlenderGym's material and geometry tasks, as opposed to the original benchmark files which use Infinigen. We use a simplified editing process, which accepts starter code and a reference image, and performs 4 rounds of sequential editing and rendering to produce a final set of renders, different to the tree of editing and verification in BlenderAlchemy \citep{huang2024blenderalchemy}. We intend only to demonstrate the relative usefulness of APIs, not the absolute performance of any LLM system --- state-of-the-art harnesses would likely be superior. See Appendix~\ref{sec:supp_llm} for details and prompts used.

\begin{table*}%
\caption{\textbf{Procedural Parameter Editing} - We use both Infinigen and ProcFunc's APIs as representations for VLMs to edit procedural generators. ProcFunc slightly improves visual metrics, despite being used in zero-shot, as opposed to Infinigen which is likely included in model training data.}
\resizebox{\textwidth}{!}{
\centering
\begin{tabular}{ll|cccc|ccccc}
\toprule
\multicolumn{2}{l|}{} & \multicolumn{4}{c|}{Materials} & \multicolumn{5}{c}{Geometry} \\
model & api & PL $\downarrow$ & NCLIP $\downarrow$ & Error \% $\downarrow$ & Cost $\downarrow$ & PL $\downarrow$ & NCLIP $\downarrow$ & CD $\downarrow$ & Error \% $\downarrow$ & Cost $\downarrow$ \\
\midrule
gemini-2.5-flash & infinigen & 4.64 & 10.69 & 4.1\% & \underline{\$5.30} & \underline{6.53} & \underline{11.77} & 2.98 & 12.80\% & \underline{\$0.98} \\
gemini-2.5-flash & procfunc & \underline{4.14} & \underline{10.18} & \underline{3.5\%} & \$6.48 & 6.87 & 12.13 & \underline{1.97} & \underline{9.40\%} & \$1.03 \\
\midrule
gemini-2.5-pro & infinigen & 8.22 & 12.43 & \underline{0.9\%} & \$7.63 & 5.98 & 8.57 & 1.65 & \underline{6.10\%} & \underline{\$3.52} \\
gemini-2.5-pro & procfunc & \underline{3.14} & \underline{7.52} & 2.0\% & \underline{\$5.72} & \underline{4.63} & \underline{7.07} & \underline{1.43} & \underline{6.10\%} & \$4.09 \\
\midrule
gpt5.2-mini & infinigen & 5.18 & 12.61 & \underline{0.0\%} & \$1.56 & \underline{6.99} & 12.19 & 2.81 & \underline{2.20\%} & \$1.21 \\
gpt5.2-mini & procfunc & \underline{3.95} & \underline{10.18} & 2.8\% & \underline{\$1.31} & 7.54 & \underline{10.54} & \underline{2.03} & 4.40\% & \underline{\$1.19} \\
\midrule
gpt5.2 & infinigen & \underline{2.93} & \underline{9.38} & \underline{0.0\%} & \$6.59 & 5.61 & 8.44 & 1.55 & \underline{0.60\%} & \underline{\$5.01} \\
gpt5.2 & procfunc & 3.93 & 9.82 & \underline{0.0\%} & \underline{\$4.97} & \underline{5.39} & \underline{8.29} & \underline{1.02} & 1.70\% & \underline{\$5.01} \\
\midrule\midrule
Averaged & infinigen & 5.24 & 11.28 & \textbf{1.3\%} & \$5.27 & 6.28 & 10.24 & 2.25 & 5.43\% & \textbf{\$2.68} \\
Averaged & procfunc & \textbf{3.79} & \textbf{9.43} & 2.1\% & \textbf{\$4.62} & \textbf{6.11} & \textbf{9.51} & \textbf{1.61} & \textbf{5.40\%} & \$2.83 \\
\bottomrule
\end{tabular}%
}
\label{tab:paramedit}
\vspace{-1.5em}
\end{table*}

We compute metrics as in BlenderGym, as well as error rate and cost. \textit{Photometric Loss} (PL) is the  mean squared error between final renders and goal images. \textit{Negative CLIP Distance} (N-CLIP) is 1 minus the cosine-similarity of feature vectors extracted by applying the \texttt{openai/clip-vit-base-patch32} pretrained ViT encoder. \textit{Chamfer Distance} is computed using 10k-point pointclouds extracted from the generated and goal geometries. \textit{PL, NCLIP and CD} are scaled by $10^2$ as in BlenderGym. We also add \textit{Error Rate}, the percentage of code editing steps which crash, and \textit{Cost}, the estimated API dollar cost of each method. 

ProcFunc slightly improves visual metrics (Tab.~\ref{tab:paramedit}) despite significant disadvantages in the standard BlenderGym setting. BlenderGym primarily focuses on editing existing numerical parameters of procedural generators, which makes adding new code and providing clear input arguments less necessary. ProcFunc is also used in zero-shot, as opposed to Infinigen's API which we believe was included in commercial model pretraining. gpt-5.2-mini often completely re-writes the tasks into Infinigen's format, despite having been instructed to make minor edits to ProcFunc code, and despite Infinigen's API not actually being available. Despite these factors, gains from ProcFunc may stem from more concise code, or from better alignment with typical Python code in pretraining. ProcFunc represents the tasks with on average 30\% reduced code characters, but this does not yet translate to cost savings due to increased model reasoning tokens or commenting when out of distribution.

\begin{wraptable}[12]{r}{0.7\textwidth}
\vspace{-1em}
\caption{\textbf{Procedural Material Creation} - We tackle BlenderGym without starter code, which requires using the APIs to add nodes and create materials from scratch. ProcFunc significantly reduces editing errors and improves visual quality and API cost.}
\footnotesize
\begin{tabular}{ll|cccc}
\toprule
model & api & PL $\downarrow$ & NCLIP $\downarrow$ & Error \% $\downarrow$ & Cost $\downarrow$ \\
\midrule
gemini-2.5-pro & infinigen & 4.37 & 10.56 & 30.3\% & \$28.60 \\
gemini-2.5-pro & procfunc & \underline{3.95} & \underline{10.14} & \underline{4.9\%} & \underline{\$15.08} \\
\midrule
gpt5.2 & infinigen & 5.09 & 11.92 & 44.7\% & \$38.81 \\
gpt5.2 & procfunc & \underline{4.49} & \underline{11.51} & \underline{21.4\%} & \underline{\$34.21} \\
\bottomrule
\end{tabular}
\label{tab:fromscratch}
\end{wraptable}

We also create a challenging new setting of creating materials from scratch by removing the starter code from BlenderGym, presented in Tab.~\ref{tab:fromscratch}. This experiment could equally be conducted with real images, but we use BlenderGym goal images as they are known to be exactly re-creatable with procedural nodes. We perform 8 rounds to compensate for increased difficulty, and additionally provide in-context documentation and 5 completed in-context examples to compensate for the lack of starter code. VLMs make overwhelming coding errors when using Infinigen (30-44\%) but much fewer when using our interface.

\begin{figure}[tb]
  \centering
  \includegraphics[width=\textwidth]{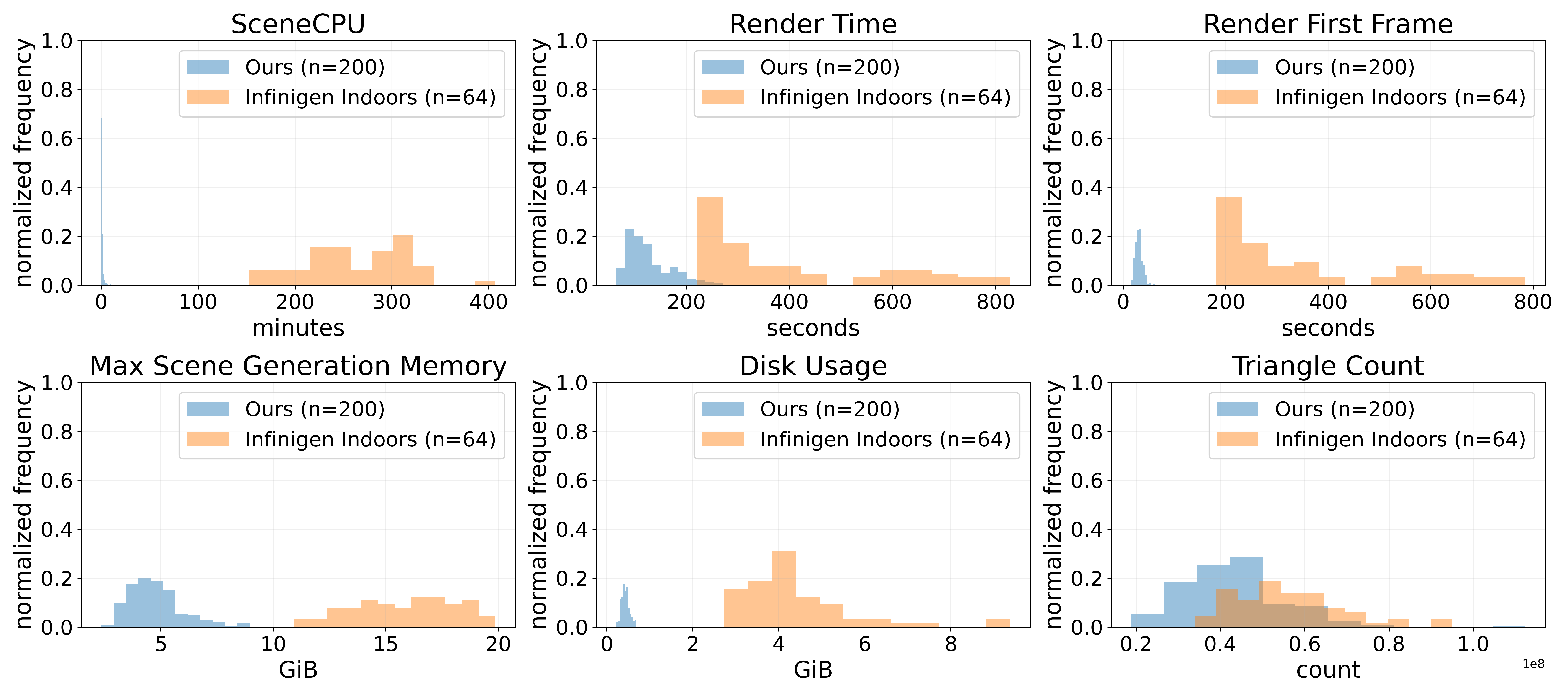}
  \caption{\textbf{Resource Usage}: Our system provides a new and useful tradeoff of performance and complexity. It matches the object count and triangle density of Infinigen-Indoors, while being significantly more efficient. Our system also has the practical benefit of more predictable resource usage.}
  \label{fig:performance_histograms}
  \vspace{-1em}
\end{figure}

\begin{table}
  \caption{\textbf{Room Generation Efficiency}. ProcFunc's included pre-implemented generators create simple room backgrounds at very low cost. Scene-CPU is the time required to generate a 3D scene prior to rendering or export. Render $t_1+t_2$ means that the first frame takes $t_1$ seconds to generate, including setup time, and the remaining frames take $t_2$ seconds per frame. }
  \label{tab:perf}
  \centering
  \resizebox{\textwidth}{!}{%
  \begin{tabular}{@{}lcc@{\hspace{0.7em}}rr@{\hspace{0.7em}}rrrrr@{}}
  \toprule
  \multicolumn{3}{c}{Settings} &
  \multicolumn{2}{c}{Complexity} &
  \multicolumn{4}{c}{Performance} \\
  \cmidrule(lr){1-3}\cmidrule(lr){4-5}\cmidrule(lr){6-9}
    Method & Obj? & Hipoly? & \#Tris & \#Objs. & SceneCPU $\downarrow$ & Mem. $\downarrow$ & Render $\downarrow$ & Storage $\downarrow$ \\
    \midrule    
    InfIndoors & $\times$ & $\times$ & 0.37M & 22.9 & 0.46min & 1.2G & \textbf{7.6$+$2.67s} & 0.35G\\
    Ours & $\times$ & $\times$ & 0.02M & 18.5 & \textbf{0.02min} & \textbf{0.41G} & 10.4$+$8.01s & 0.16G \\
    Ours Eevee & $\times$ & $\times$ & 0.02M & 18.5 & \textbf{0.02min} & \textbf{0.41G} & 71.9$+$4.44s & \textbf{0.15G} \\
    \addlinespace
    InfIndoors & $\times$ & $\checkmark$ & 49.1M & 22.7 & 276min & 15.07G & 938.1$+$3.82s & 3.78G \\
    Ours & $\times$ & $\checkmark$ & 35.4M & 18.5 & \textbf{0.02min} & \textbf{2.53G} & \textbf{25.8$+$8.49s} & \textbf{0.16G} \\
    \addlinespace
    InfIndoors & $\checkmark$ & $\times$ & 6.92M & 31.9 & 17.91min & \textbf{2.67G} & 196.1$+$\textbf{3.27s}& 0.77G \\
    Ours & $\checkmark$ & $\times$ & 3.13M & 60.1 & 0.77min & 4.11G & \textbf{13.1}$+$8.19s & 0.42G \\
    Ours Eevee & $\checkmark$ & $\times$ & 3.22M & 59.3 & \textbf{0.69min} & 4.11G & 105.7$+$5.46s & \textbf{0.41G} \\
    \addlinespace
    InfIndoors & $\checkmark$ & $\checkmark$ & 55.7M & 30.6 & 278.98min & 15.94G & 356.6$+$\textbf{3.66s} & 4.39G \\
    Ours & $\checkmark$ & $\checkmark$ & 43.4M & 60.1 & \textbf{1.1min} & \textbf{4.743G} & \textbf{30.7}$+$8.79s & \textbf{0.43G} \\
    \bottomrule
  \end{tabular}%
  }
  \vspace{-1.5em}
\end{table}

\vspace{-0.6em}

\subsection{Procedural Generator Efficiency}

We aim to evaluate the resource-efficiency of the example procedural room generator (Sec.~\ref{sec:basegen}). Efficiency is generally a tradeoff with scene detail and complexity: higher triangle-count or object-count scenes require both more memory and render time on the GPU. Thus, we evaluate both high-detail and low-detail scenes, and report both the efficiency and complexity of each result. 

We benchmark our system by generating many scenes and rendering 12 monocular camera views in each scene. We report hardware usage (CPU time, RAM, GPU time, Storage) so that individuals can determine how this maps to throughput or monetary cost. SceneCPU refers to the CPU-only duration to create a 3D scene file (which can be used for any purpose), and assumes access to 4 Intel(R) Xeon(R) Gold 5320 CPU cores. GPU time is measured on 1 Nvidia L40 GPU. 

We provide results in Tab.~\ref{tab:perf} and Fig.~\ref{fig:performance_histograms}. Our system has very short CPU runtime (less than 2 minutes, even for detailed scenes). Render time is comparable to or slightly slower than Infinigen Indoors, due to the cost of executing more detailed shader programs. This could be reduced by pre-evaluating shaders to image textures. We also reduce scene and groundtruth storage cost by 40\% to 90\%.

\subsection{Procedural Generator Diversity} Composition increases the semantic coverage of any given set of procedural generators. Without composition each attribute is present by itself, whereas with composition, all combinations are present without being implemented one by one. This is best measured within a single system, with the set of attributes held fixed, rather than between systems with different sets or granularity of attributes. We use our program tracer to extract compute graphs with and without composition and analytically calculate their diversity (Tab.~\ref{tab:diversity},Fig.~\ref{fig:diversity}). Entropy and other statistics are meaningful only if all procedural generator parameters produce useful visual changes, so we verify this is true to the best of our ability.

\begin{table}
  \caption{\textbf{Analytical Diversity} -  Composition increases the mean number of discrete choices and continuous parameters per generator, increases the entropy of the distribution (assuming continuous parameters discretized to 3 bits), and increases Cyclomatic / McCabe complexity, which measures the number of linearly independent control flow paths.}
  \vspace{0.2em}
  \label{tab:diversity}
  \centering
  \resizebox{0.8\textwidth}{!}{%
  \begin{tabular}{@{}l@{\hspace{1.2em}}c@{\hspace{1.2em}}c@{\hspace{1.2em}}c@{\hspace{1.2em}}c@{}}
   \toprule
    Strategy & \shortstack{Continuous Params.} & \shortstack{Discrete Params.} & Cyclomatic Complexity & Entropy (bits)\\
    \midrule
    Base Materials & 25.5 & 1.80 & 45 & 81.3\\
    Multi-level Composition & 41.7 & 3.71 & 448 & 134\\
  \bottomrule
  \end{tabular}
  }
\end{table}

\subsection{Procedural Material Geometry Detail} 

\begin{wrapfigure}[11]{tr}{0.4\textwidth}
    \vspace{-3em}
    \centering
    \includegraphics[width=0.4\textwidth]{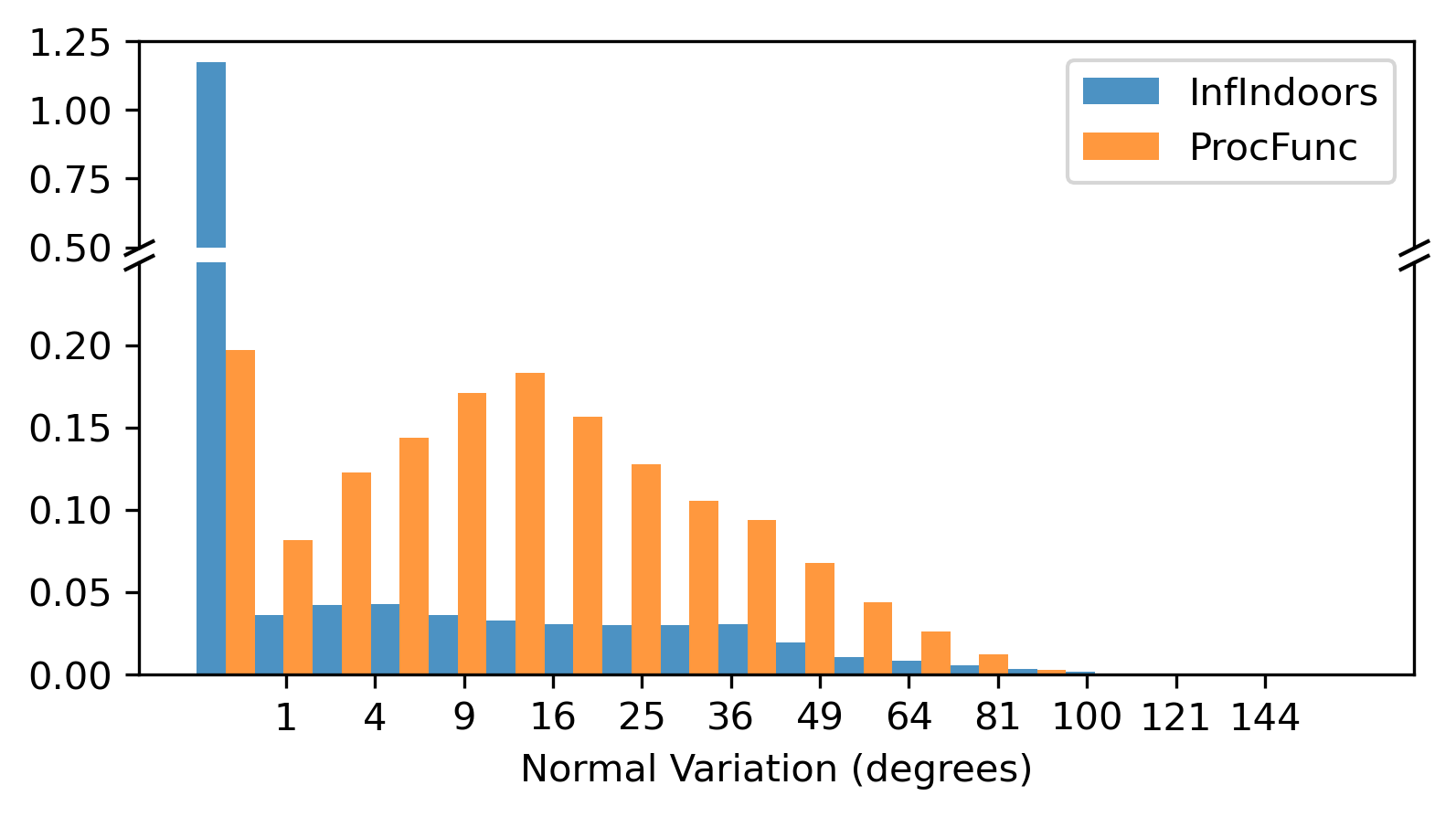}
    \vspace{-1.5em}
    \caption{\textbf{Surface Detail} - Our room materials have significantly more varied surface normals than Infinigen Indoors.}
    \label{fig:gt_distrib}
\end{wrapfigure}

ProcFunc's provided material generators are designed to create detailed meshes by compositionally combining per-vertex displacement shaders. Besides providing excellent visual realism (Fig.~\ref{fig:livingrooms}), this provides very detailed ground truth annotations for tasks such as depth estimation or surface normal estimation. In Fig.~\ref{fig:gt_distrib} we compute the detailedness of these annotations via average normal variation $V_i = \sum_{j\in N(i)} \angle(n_i, n_j)$ for square neighborhoods of size $15 \times 15$ centered at pixel $i$. We aggregate valid pixels from 500 images in each dataset.

\begin{table}
  \caption{\textbf{Stereo Dataset Comparison}: Our system is able to generate highly effective stereo data. Midd (F) and (H) refer to full and half resolution of the Middlebury 2014 validation set, respectively. We evaluate on  quarter resolution Booster.}
  \label{tab:stereo}
  \centering
  \setlength{\tabcolsep}{4pt}
  \begin{tabular}{@{}l r%
    r@{\hspace{0.6em}}r @{\hspace{1.0em}}%
    r@{\hspace{0.6em}}r @{\hspace{1.0em}}%
    r@{\hspace{0.6em}}r @{\hspace{1.0em}}%
    r@{\hspace{0.6em}}r @{\hspace{1.0em}}%
    r@{\hspace{0.6em}}r@{}}
    \toprule
    \multicolumn{1}{c}{Dataset} &
    \multicolumn{1}{c}{Size} &
    \multicolumn{2}{c}{Midd (F)} &
    \multicolumn{2}{c}{Midd (H)} &
    \multicolumn{2}{c}{KITTI-15} &
    \multicolumn{2}{c}{ETH3D} &
    \multicolumn{2}{c}{Booster} \\
    \cmidrule(lr){3-4}\cmidrule(lr){5-6}\cmidrule(lr){7-8}\cmidrule(lr){9-10}\cmidrule(lr){11-12}
    & & EPE & 2px & EPE & 2px & EPE & 3px & EPE & 1px & EPE & 2px \\
    \midrule
    Sceneflow  & 40k  & 4.37 & 13.72 & 1.71 & 9.27 & 2.59 & 6.03 & 0.30 & 3.83 & 3.77 & 15.34 \\
    CREStereo  & 200k & 5.28 & 17.24 & 2.09 & 13.81 & 1.19 & 6.27 & \textbf{0.28} & 3.85 & 2.67 & 14.17 \\
    TartanAir  & 306k & 10.7 & 18.24 & 2.17 & 8.92 & 1.09 & 5.49 & 0.35 & 4.3  & 3.89 & 14.01 \\
    FSD        & 1M  & \textbf{1.67} & \textbf{7.79}  & \textbf{0.83} & \textbf{5.93} & \textbf{1.03} & \underline{5.00} & 0.29 & \underline{2.98} & \underline{2.04} & \underline{10.04} \\
    ProcFunc       & 50k  & \underline{2.23} & \underline{9.05} & \underline{1.16} & \underline{6.86} & \underline{1.06} & \textbf{4.92} & \textbf{0.28} & \textbf{2.95} & \textbf{1.42} & \textbf{9.23}  \\
    \bottomrule
  \end{tabular}

\end{table}

\subsection{Stereo Dataset Generation}
\label{sec:stereo}
We provide a baseline result for the effectiveness of stereo datasets generated using our minimal base room generator. We generate a dataset of 50k stereo pairs using our room generator combined with a small subset of design choices from WMGStereo \citep{yan2026makesgoodsynthetictraining}, namely random floating objects and removing reflective materials. Compared to the WMGStereo scene generator, which builds on an optimized version of Infinigen Indoors, our stereo scenes were 31x faster to generate (0.41 min vs. 13 min).

We train RAFT-Stereo \citep{raftstereo} on a shorter training schedule of 75k steps using standard hyper-parameters and data augmentation using 2 NVIDIA A6000 GPUs. We provide results on standard benchmarks in Tab.~\ref{tab:stereo}. Training on ProcFunc data yielded better performance than several widely used stereo datasets, and is competitive with FSD's dataset \citep{fsd}, despite being drastically smaller. Notably, we do not claim an advance in state-of-the-art stereo dataset generation. Rather, we show that a simple ProcFunc program serves as a strong baseline for stereo data. 

\section*{Acknowledgements}
This work was partially supported by the National Science Foundation.

\bibliographystyle{plainnat}
\bibliography{main}

\newpage
\appendix
\newcommand{\parbf}[1]{\par \noindent \textbf{#1}.}
\newcommand{\parit}[1]{\par \noindent \textit{#1}.}

\renewcommand{\thesection}{\Alph{section}}

\lstdefinestyle{prompt}{
    basicstyle=\ttfamily\small,
    breaklines=true,
    breakatwhitespace=false,
    frame=single,
    rulecolor=\color{black!30},
    backgroundcolor=\color{gray!5},
    xleftmargin=0.5em,
    xrightmargin=0.5em,
    aboveskip=1em,
    belowskip=1em,
    showstringspaces=false,
    columns=fullflexible,
    keepspaces=true,
    firstline=4,  %
}

\lstdefinestyle{pythonstyle}{
    language=Python,
    basicstyle=\ttfamily\scriptsize,
    keywordstyle=\color{blue!70!black}\bfseries,
    stringstyle=\color{orange!80!black},
    commentstyle=\color{gray}\itshape,
    identifierstyle=\color{black},
    numberstyle=\tiny\color{gray},
    numbers=left,
    numbersep=4pt,
    stepnumber=5,
    showstringspaces=false,
    breaklines=true,
    breakatwhitespace=true,
    frame=single,
    framesep=3pt,
    xleftmargin=10pt,
    tabsize=4,
    columns=flexible,
    keepspaces=true,
    morekeywords={def, return, None, True, False},
}

\section{Qualitative Results}
\label{sec:supp_qual}

\subsection{Room Generation Random Sample}
\label{sec:supp_rooms}
\begin{figure}[th]
  \centering
  \includegraphics[width=\textwidth]{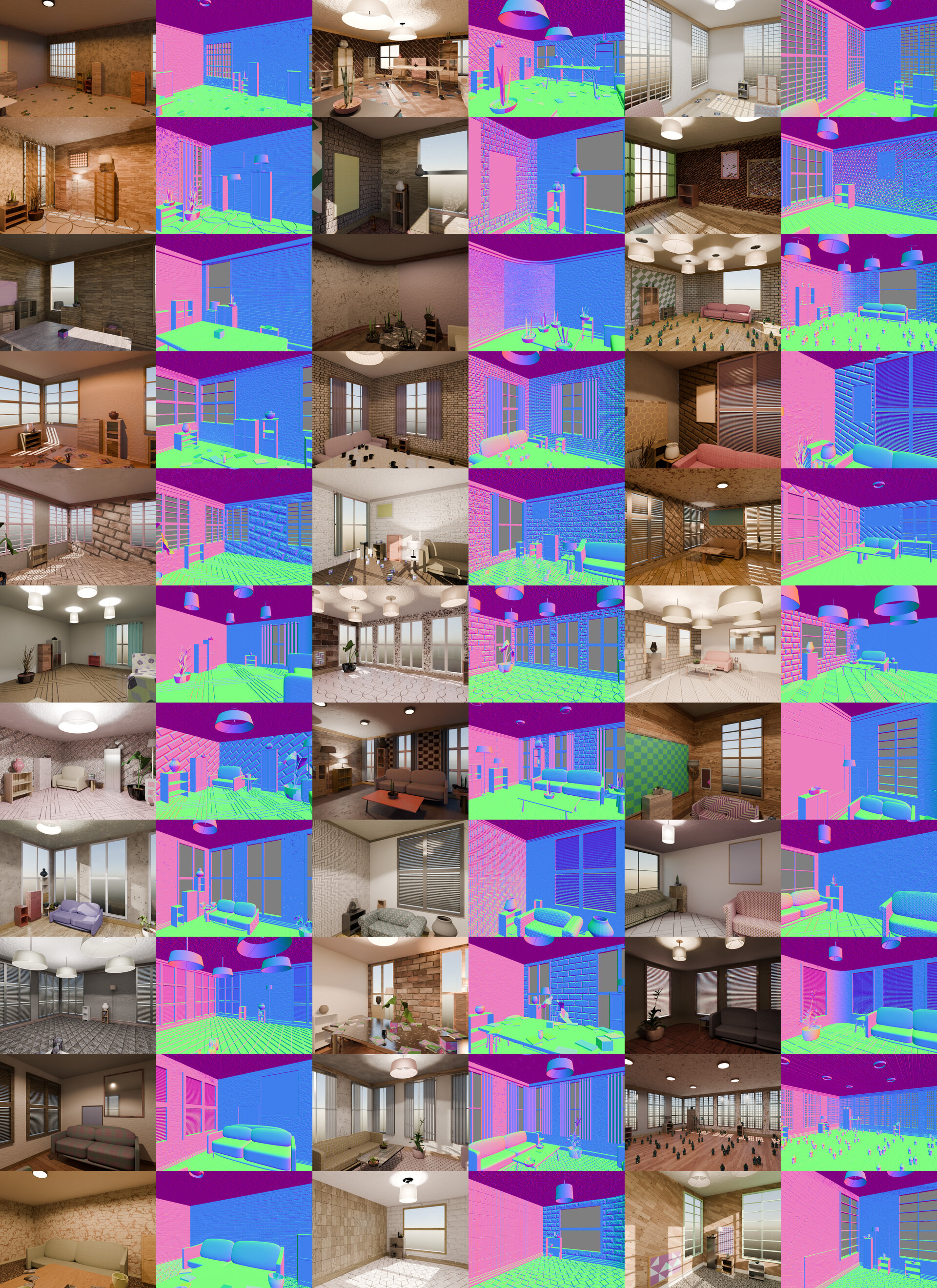}

  \caption{Sample living rooms from our procedural scene generator paired with surface normal ground-truth 1/2.}
  \label{fig:livingroom_normals_1}
\end{figure}

\begin{figure}[th]
  \centering
  \includegraphics[width=\textwidth]{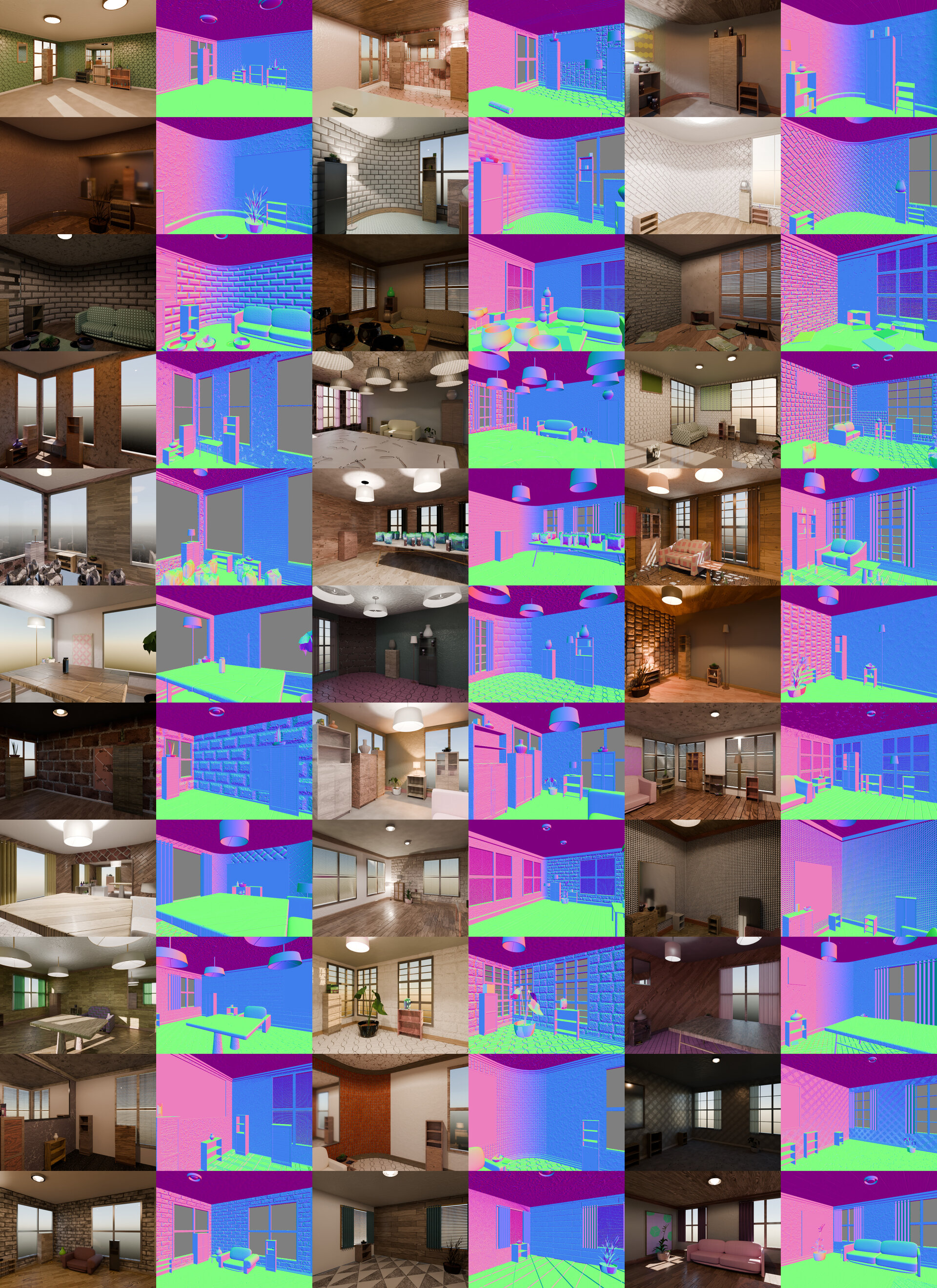}

  \caption{Sample living rooms from our procedural scene generator paired with surface normal ground-truth 2/2.}
  \label{fig:livingroom_normals_2}
\end{figure}

\cref{fig:livingroom_normals_1} and \cref{fig:livingroom_normals_2} show random livingroom samples from our procedural scene generator paired with surface-normal ground truth. 

\subsection{Stereo Dataset Random Sample}
\label{sec:supp_stereo}

\begin{figure*}[p]                                                             
  \centering      
  \includegraphics[width=\textwidth]{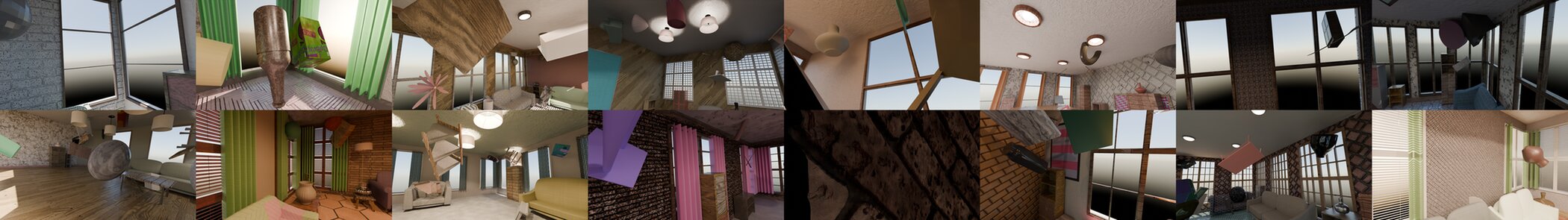}\\[2pt]
  \includegraphics[width=\textwidth]{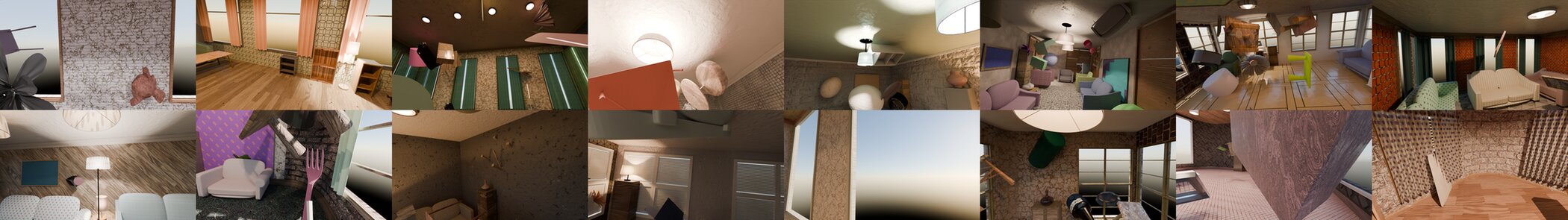}\\[2pt]
  \includegraphics[width=\textwidth]{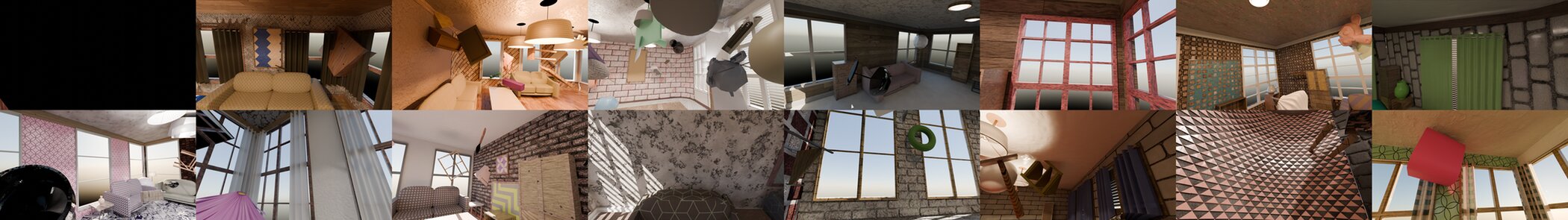}\\[2pt]
  \includegraphics[width=\textwidth]{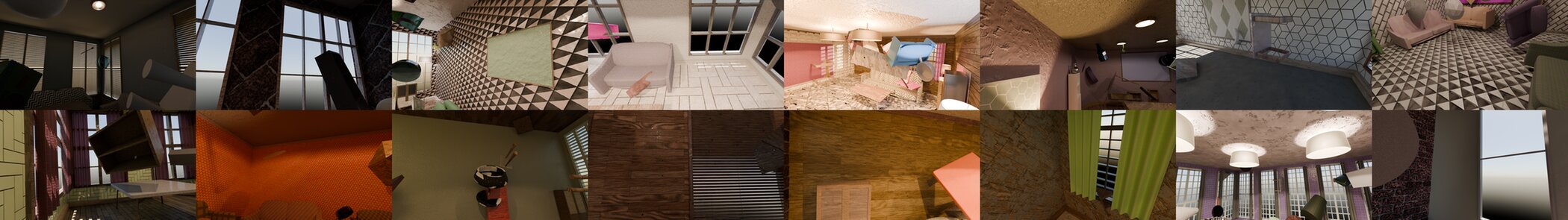}\\[2pt]
  \includegraphics[width=\textwidth]{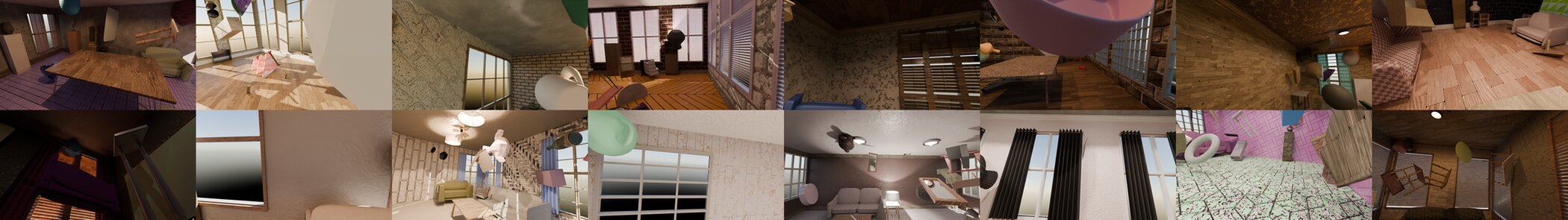}\\[2pt]
  \includegraphics[width=\textwidth]{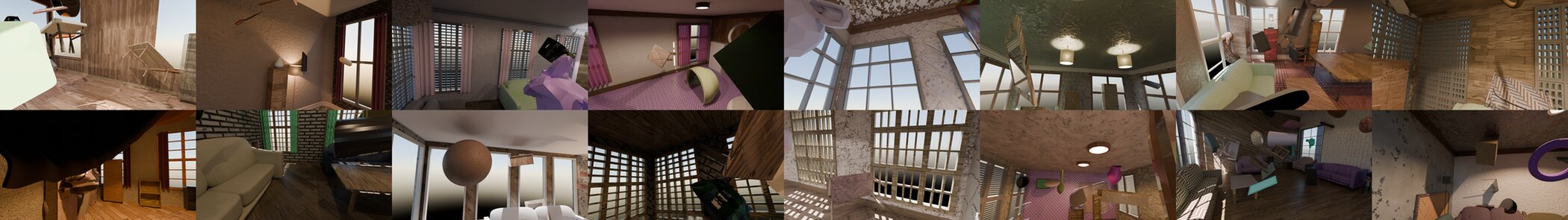}\\[2pt]
  \includegraphics[width=\textwidth]{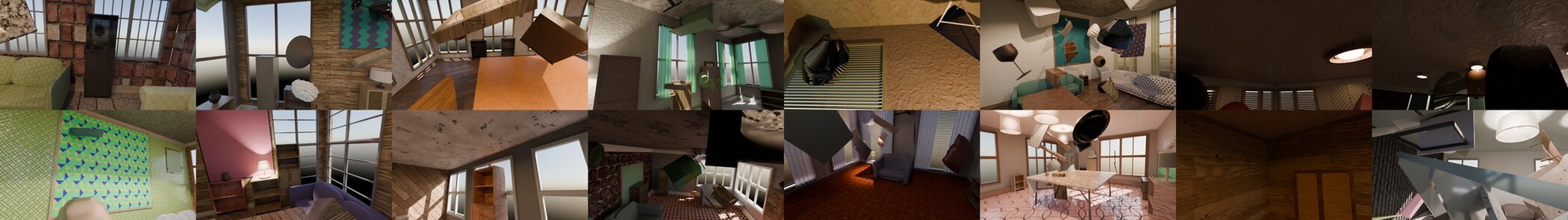}\\[2pt]
  \includegraphics[width=\textwidth]{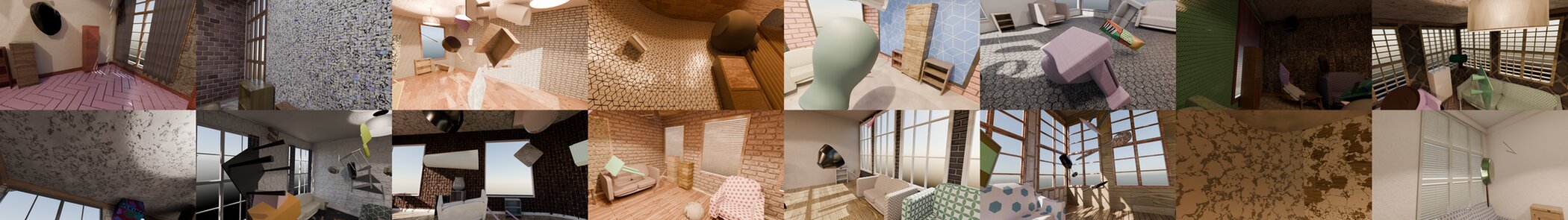}
  \caption{Raw non-preprocessed stereo dataset samples. We show one random image from each of a random subset of scenes in the dataset. In practice, we render stereo images from 12 unique camera views per scene. We also apply filtering before using them as training data, to remove cases with objects too close to the camera.}
  \label{fig:stereo_examples}
  \end{figure*}

We provide a random sample of left-camera images from our stereo training dataset in Fig.~\ref{fig:stereo_examples}

\subsection{High-Realism Materials and Material Composites}
\label{sec:supp_materials}

\begin{figure*}[h!]
    \centering
    \vspace{-2mm}
    \includegraphics[width=\textwidth]{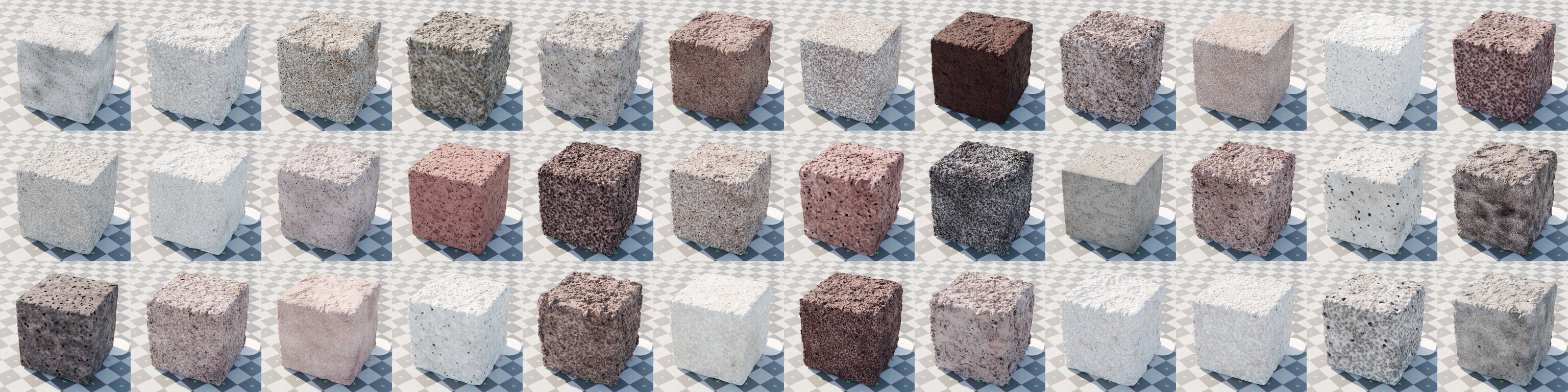}
    \vspace{-5mm}
    \caption{Brick Concrete (\texttt{brick\_concrete\_distribution})}
    \vspace{-4mm}
    \label{fig:qual_realistic_firstmat}
\end{figure*}

\begin{figure*}[h!]
    \centering
    \vspace{-2mm}
    \includegraphics[width=\textwidth]{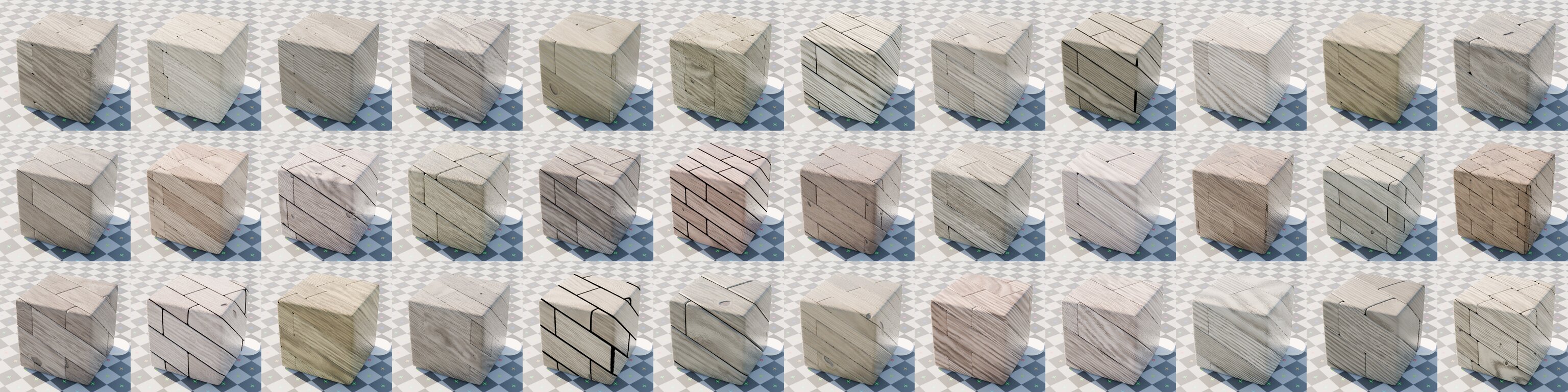}
    \vspace{-5mm}
    \caption{Wood Planks (\texttt{wood\_planks\_distribution})}
    \vspace{-4mm}
    \label{fig:qual_realistic_lastmat}
\end{figure*}

We provide 36 random samples for each high-level pre-made material generator in Figures.~\ref{fig:qual_realistic_firstmat}-\ref{fig:qual_realistic_lastmat}.

\begin{figure*}[h!]
    \centering
    \vspace{-2mm}
    \includegraphics[width=\textwidth]{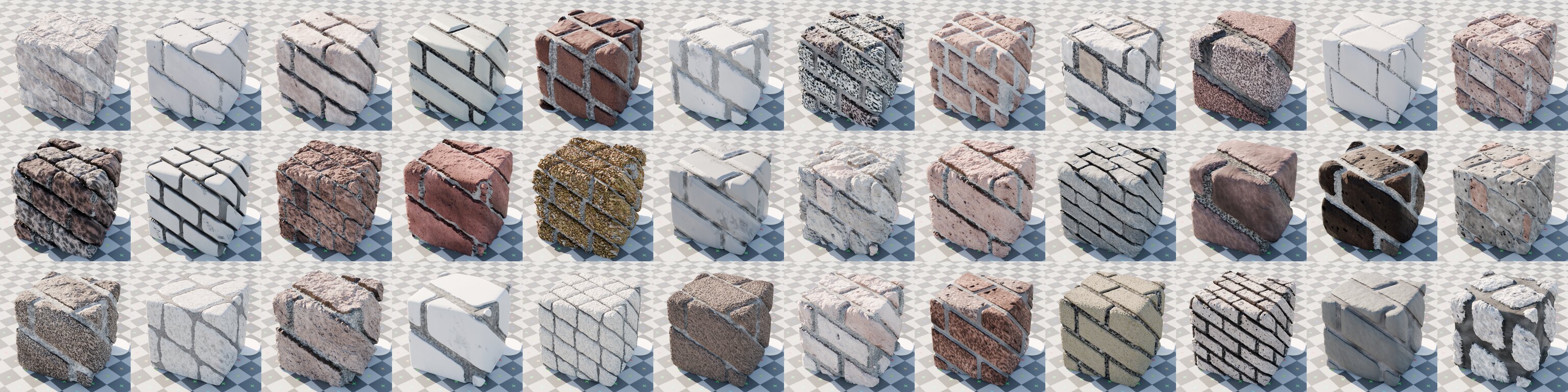}
    \vspace{-5mm}
    \caption{Bricks (\texttt{bricks\_distribution})}
    \vspace{-4mm}
\end{figure*}
\begin{figure*}[h!]
    \centering
    \vspace{-2mm}
    \includegraphics[width=\textwidth]{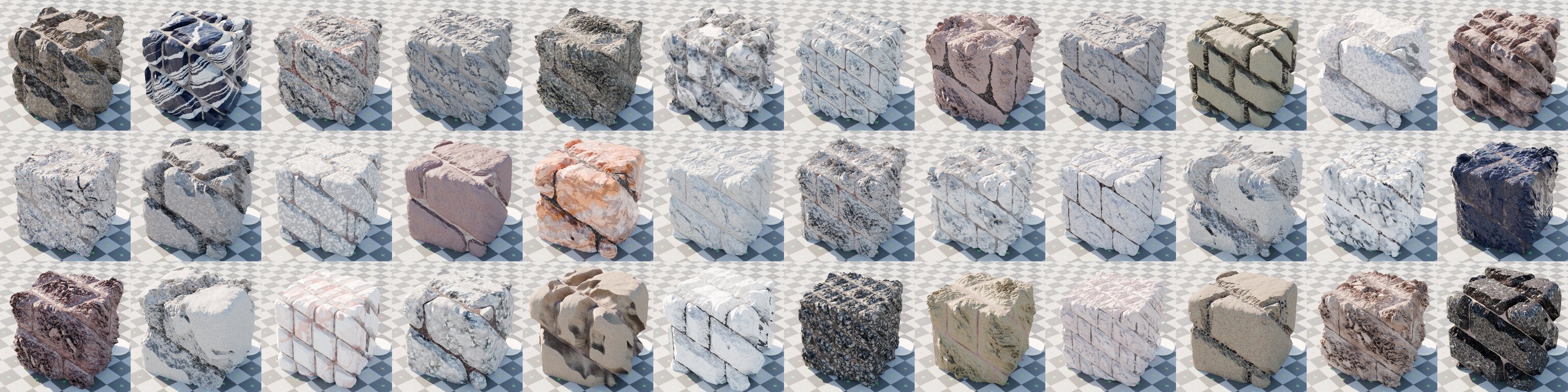}
    \vspace{-5mm}
    \caption{Bricks Masonry (\texttt{bricks\_masonry\_distribution})}
    \vspace{-4mm}
\end{figure*}
\begin{figure*}[h!]
    \centering
    \vspace{-2mm}
    \includegraphics[width=\textwidth]{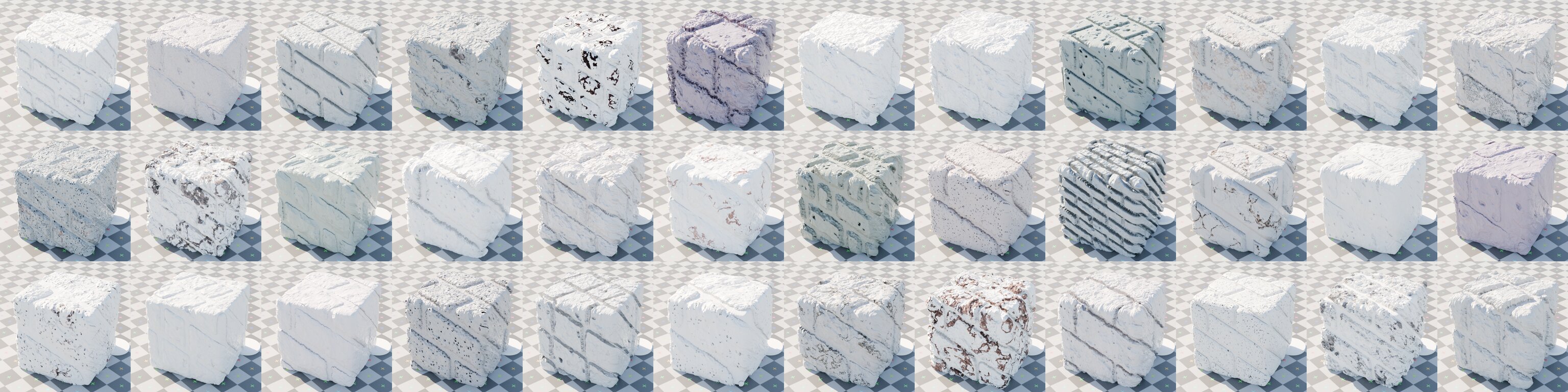}
    \vspace{-5mm}
    \caption{Bricks Paint (\texttt{bricks\_paint\_distribution})}
    \vspace{-4mm}
\end{figure*}
\begin{figure*}[h!]
    \centering
    \vspace{-2mm}
    \includegraphics[width=\textwidth]{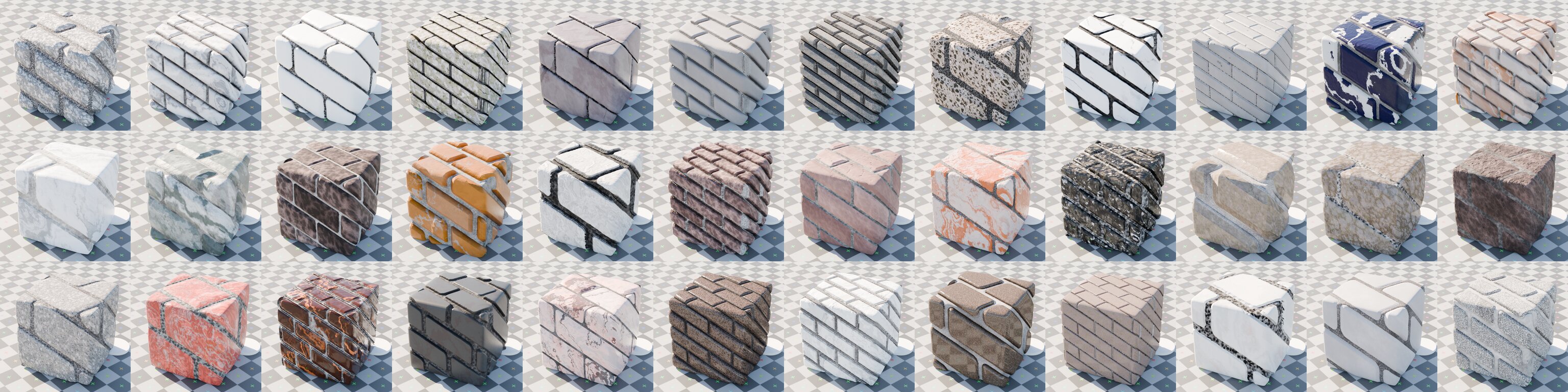}
    \vspace{-5mm}
    \caption{Bricks Pristine (\texttt{bricks\_pristine\_distribution})}
    \vspace{-4mm}
\end{figure*}
\begin{figure*}[h!]
    \centering
    \vspace{-2mm}
    \includegraphics[width=\textwidth]{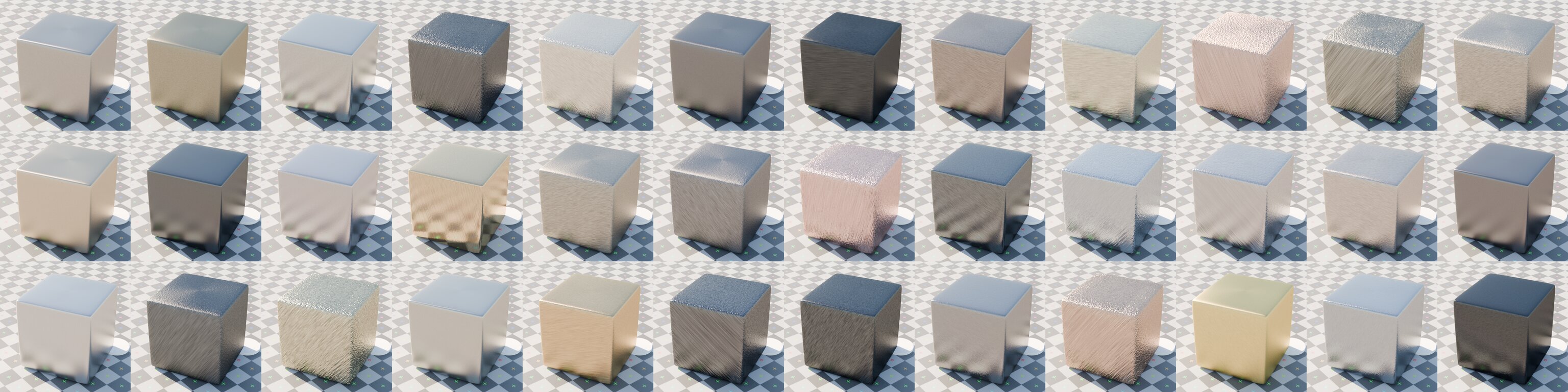}
    \vspace{-5mm}
    \caption{Brushed Metal Linear (\texttt{brushed\_metal\_linear\_distribution})}
    \vspace{-4mm}
\end{figure*}
\begin{figure*}[h!]
    \centering
    \vspace{-2mm}
    \includegraphics[width=\textwidth]{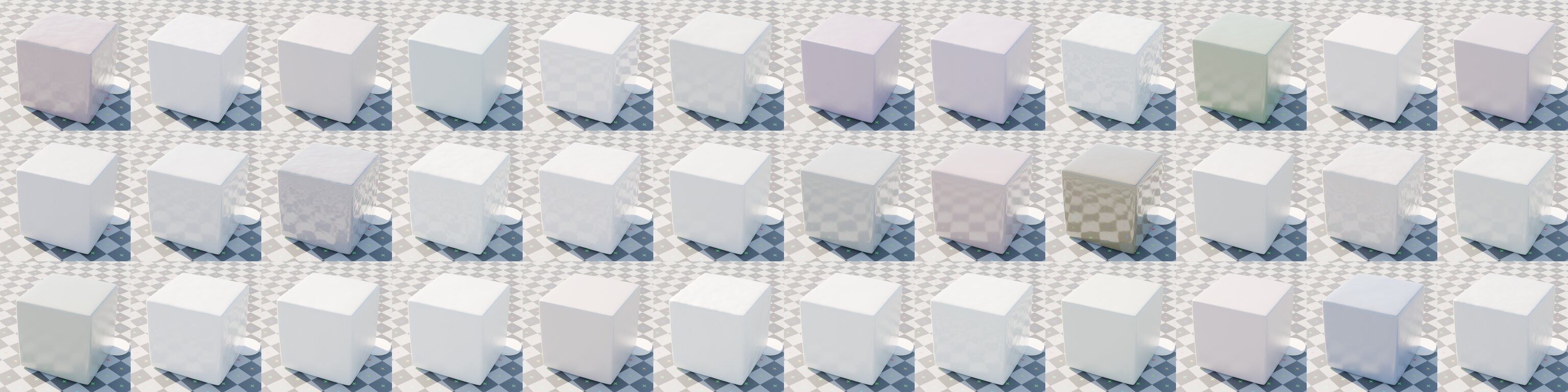}
    \vspace{-5mm}
    \caption{Ceramic (\texttt{ceramic\_distribution})}
    \vspace{-4mm}
\end{figure*}
\begin{figure*}[h!]
    \centering
    \vspace{-2mm}
    \includegraphics[width=\textwidth]{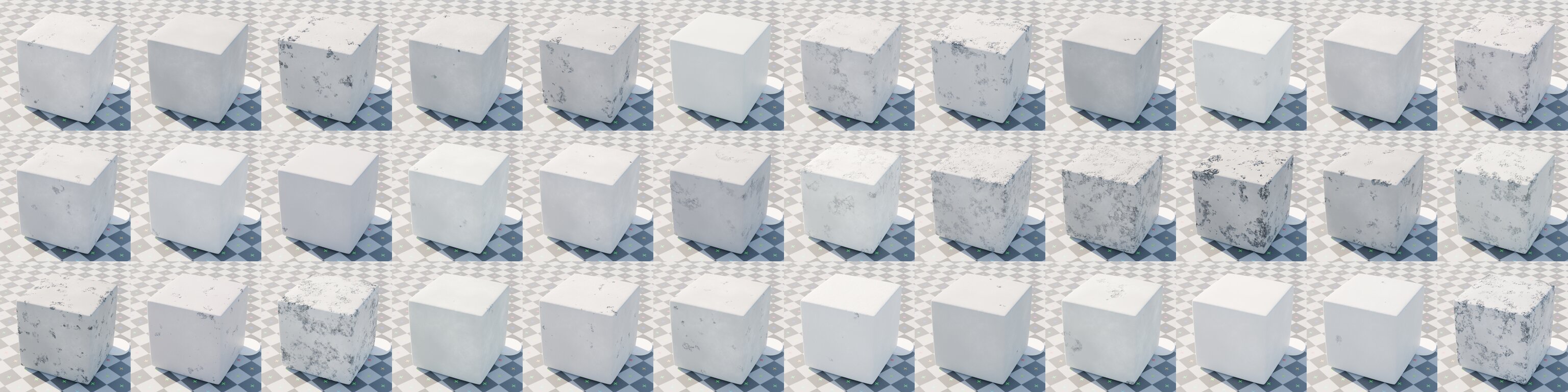}
    \vspace{-5mm}
    \caption{Concrete (\texttt{concrete\_distribution})}
    \vspace{-4mm}
\end{figure*}
\begin{figure*}[h!]
    \centering
    \vspace{-2mm}
    \includegraphics[width=\textwidth]{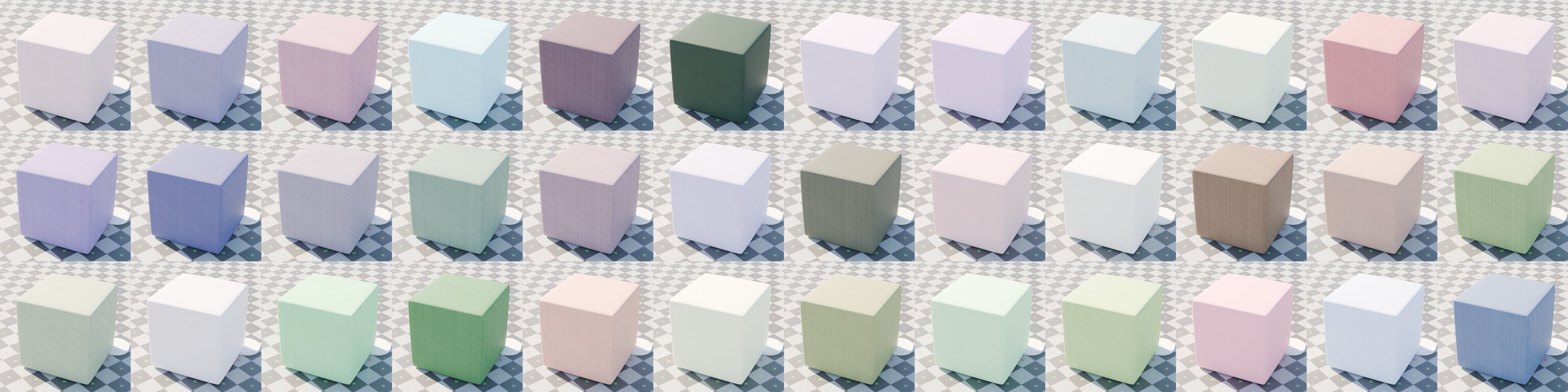}
    \vspace{-5mm}
    \caption{Fabric (\texttt{fabric\_distribution})}
    \vspace{-4mm}
\end{figure*}
\begin{figure*}[h!]
    \centering
    \vspace{-2mm}
    \includegraphics[width=\textwidth]{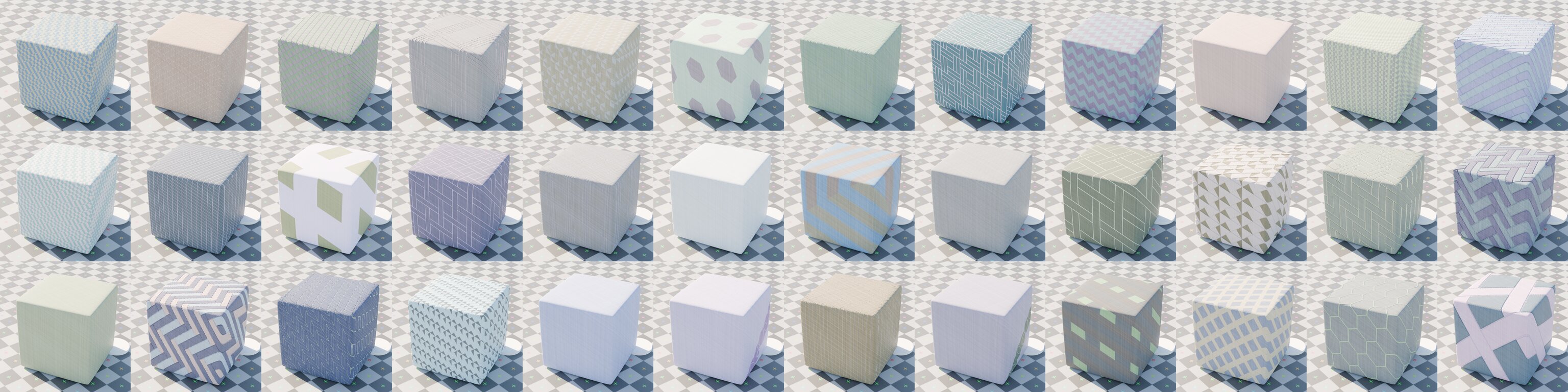}
    \vspace{-5mm}
    \caption{Fabric Patterned (\texttt{fabric\_patterned\_distribution})}
    \vspace{-4mm}
\end{figure*}
\begin{figure*}[h!]
    \centering
    \vspace{-2mm}
    \includegraphics[width=\textwidth]{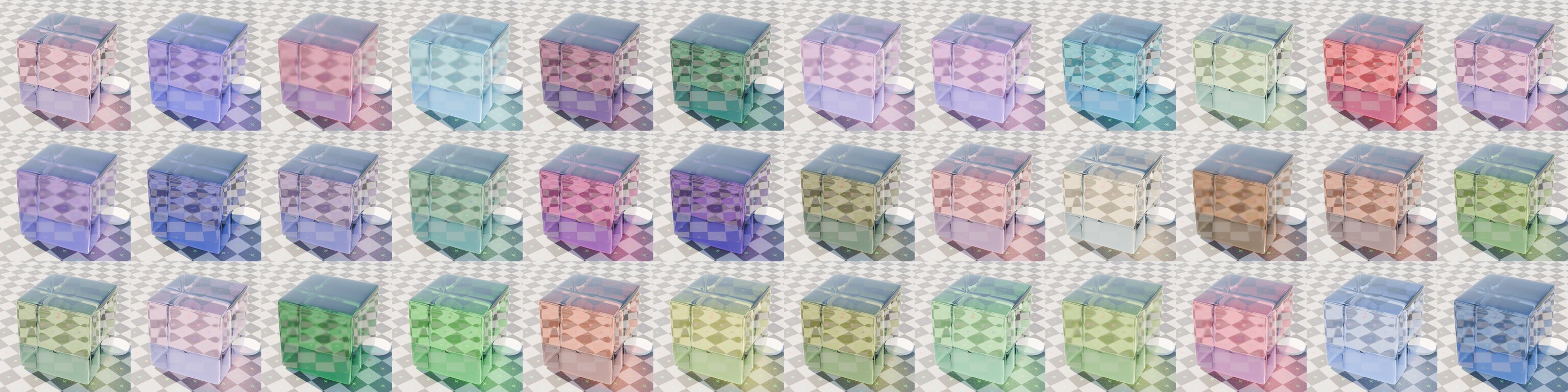}
    \vspace{-5mm}
    \caption{Glass Colored (\texttt{glass\_colored\_distribution})}
    \vspace{-4mm}
\end{figure*}
\begin{figure*}[h!]
    \centering
    \vspace{-2mm}
    \includegraphics[width=\textwidth]{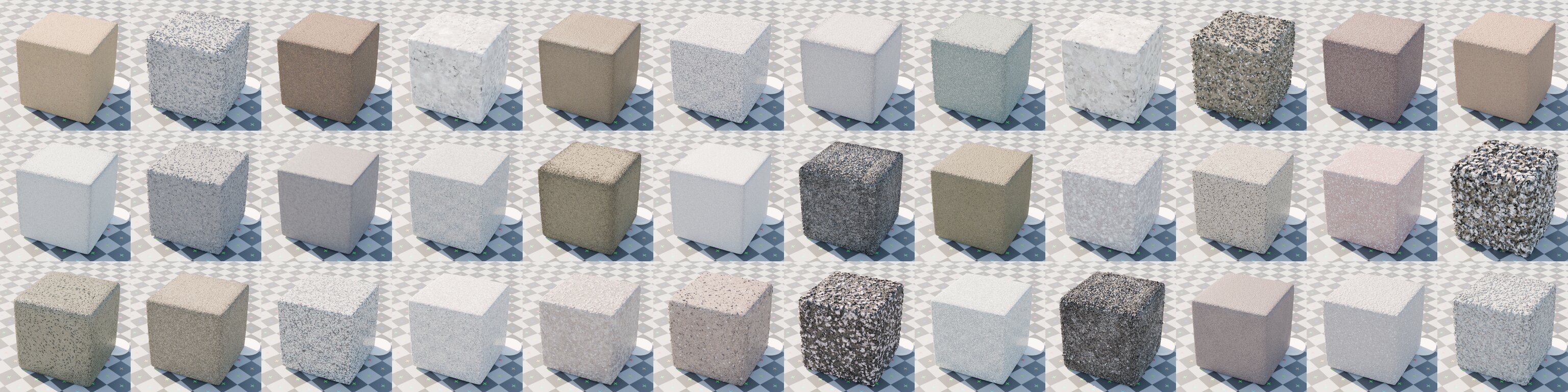}
    \vspace{-5mm}
    \caption{Granite (\texttt{granite\_distribution})}
    \vspace{-4mm}
\end{figure*}
\begin{figure*}[h!]
    \centering
    \vspace{-2mm}
    \includegraphics[width=\textwidth]{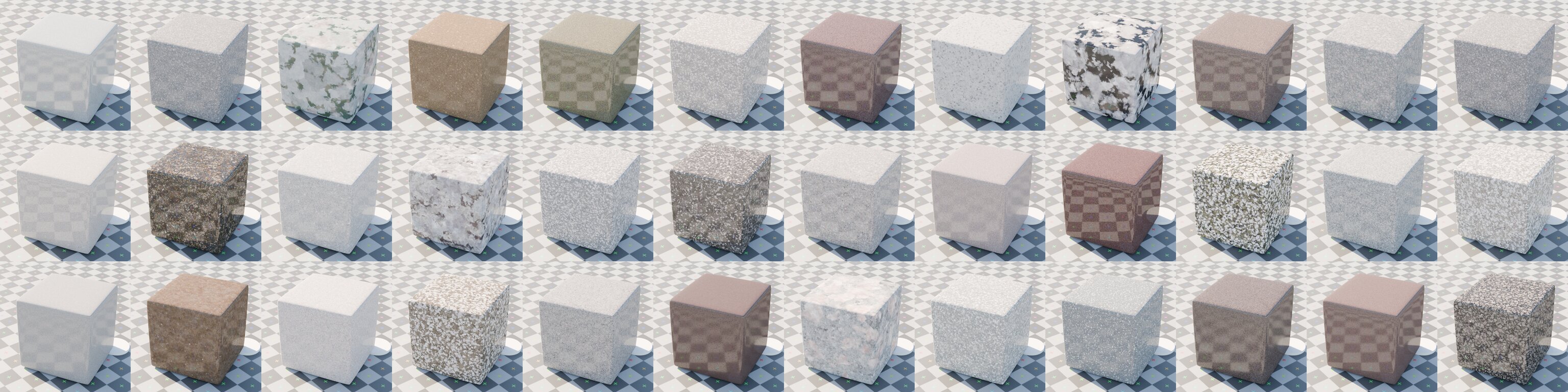}
    \vspace{-5mm}
    \caption{Granite Smooth (\texttt{granite\_smooth\_distribution})}
    \vspace{-4mm}
\end{figure*}

\begin{figure*}[h!]
    \centering
    \vspace{-2mm}
    \includegraphics[width=\textwidth]{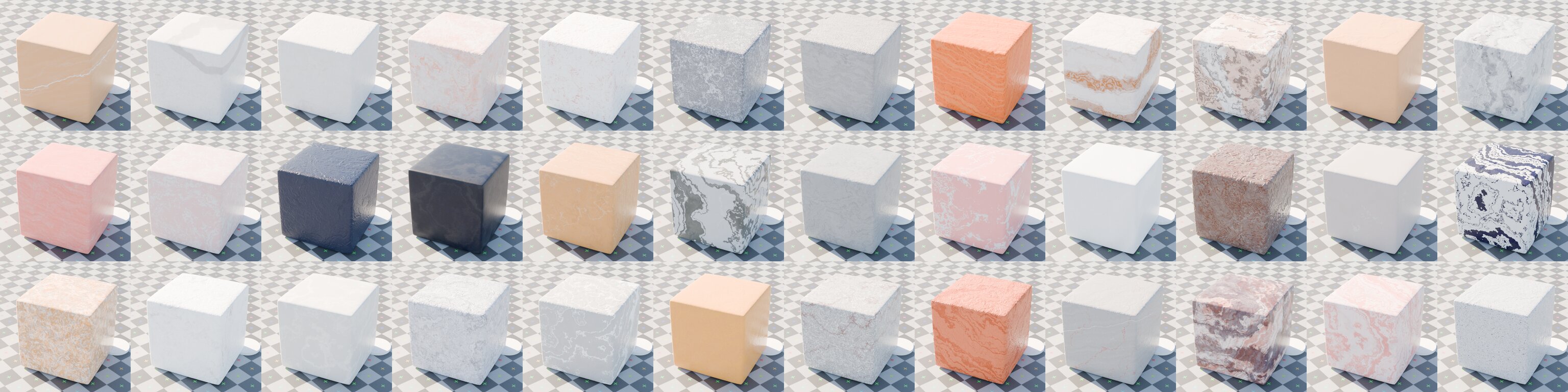}
    \vspace{-5mm}
    \caption{Marble (\texttt{marble\_distribution})}
    \vspace{-4mm}
\end{figure*}
\begin{figure*}[h!]
    \centering
    \vspace{-2mm}
    \includegraphics[width=\textwidth]{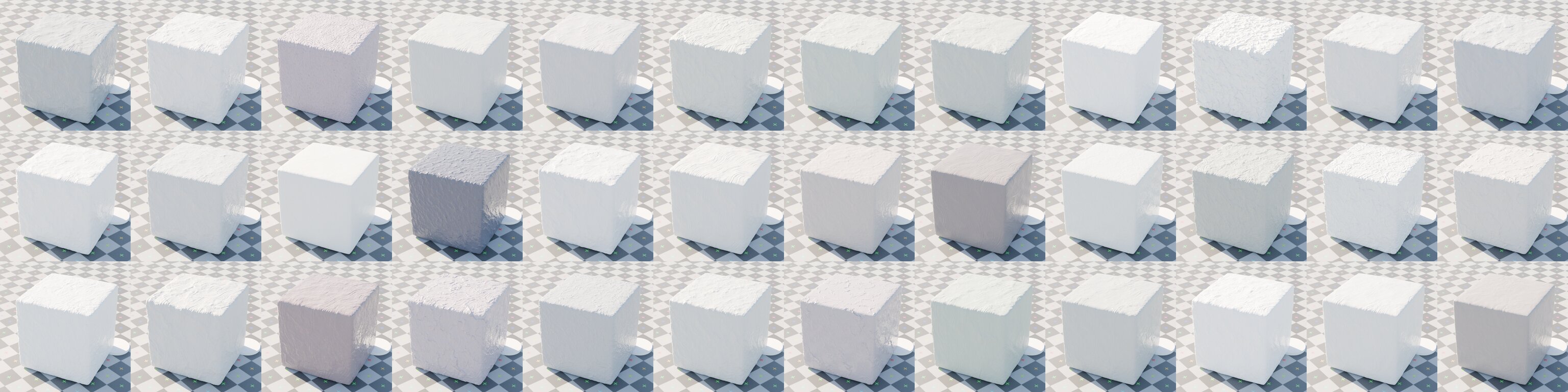}
    \vspace{-5mm}
    \caption{Paint (\texttt{paint\_distribution})}
    \vspace{-4mm}
\end{figure*}
\begin{figure*}[h!]
    \centering
    \vspace{-2mm}
    \includegraphics[width=\textwidth]{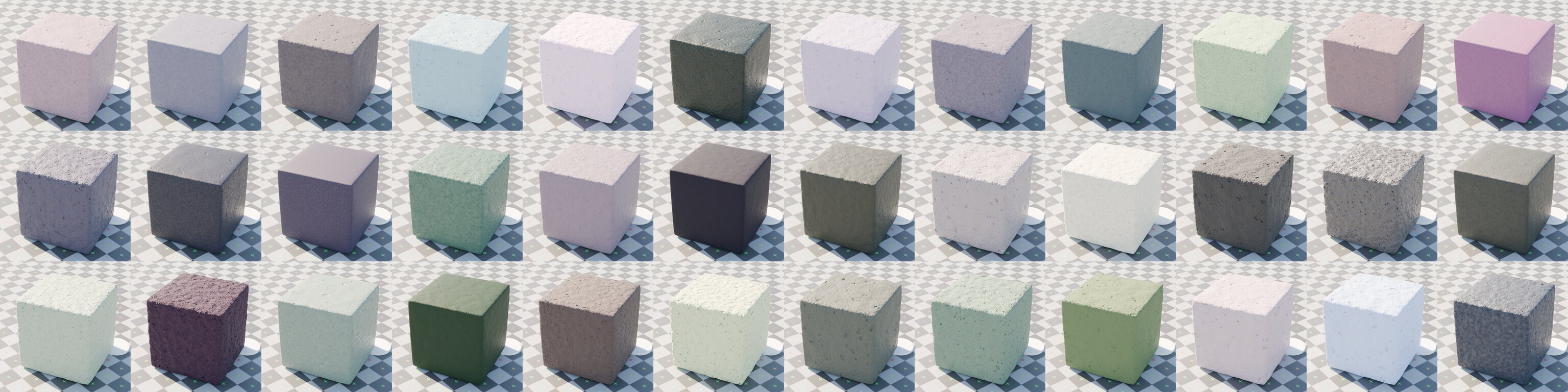}
    \vspace{-5mm}
    \caption{Plastic Opaque (\texttt{plastic\_opaque\_distribution})}
    \vspace{-4mm}
\end{figure*}
\begin{figure*}[h!]
    \centering
    \vspace{-2mm}
    \includegraphics[width=\textwidth]{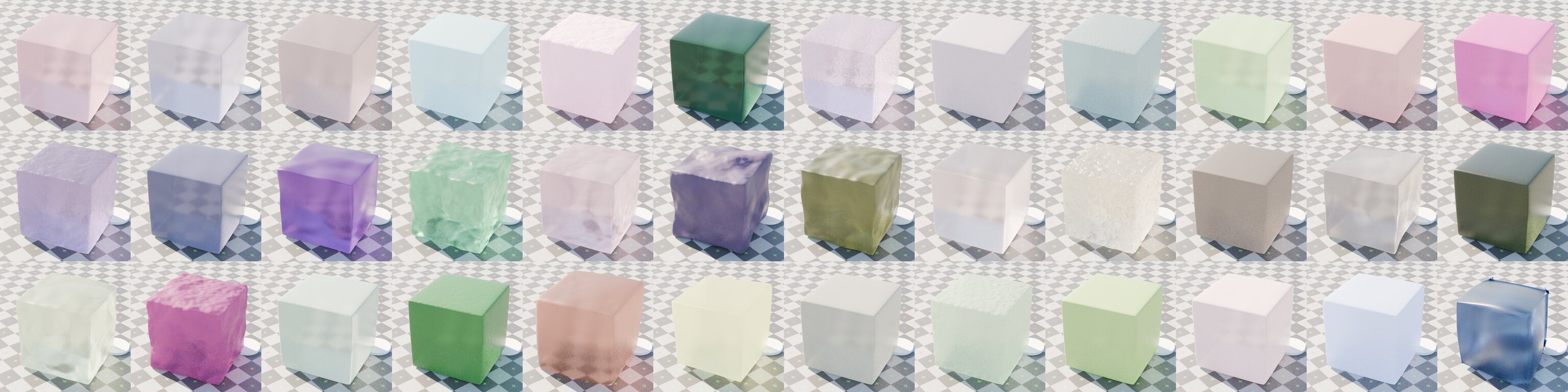}
    \vspace{-5mm}
    \caption{Plastic Translucent (\texttt{plastic\_translucent\_distribution})}
    \vspace{-4mm}
\end{figure*}
\begin{figure*}[h!]
    \centering
    \vspace{-2mm}
    \includegraphics[width=\textwidth]{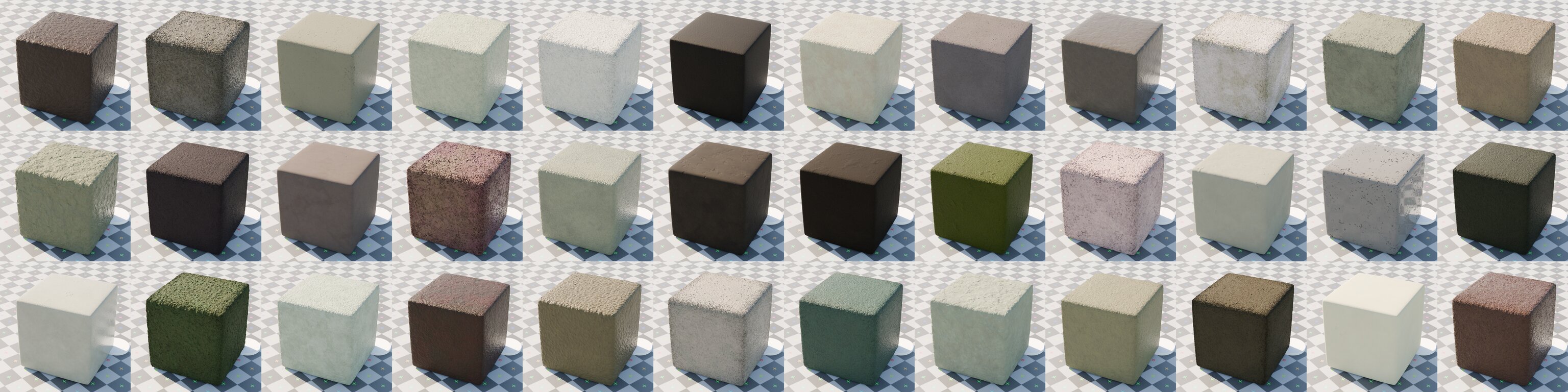}
    \vspace{-5mm}
    \caption{Stone Smooth (\texttt{stone\_smooth\_distribution})}
    \vspace{-4mm}
\end{figure*}
\begin{figure*}[h!]
    \centering
    \vspace{-2mm}
    \includegraphics[width=\textwidth]{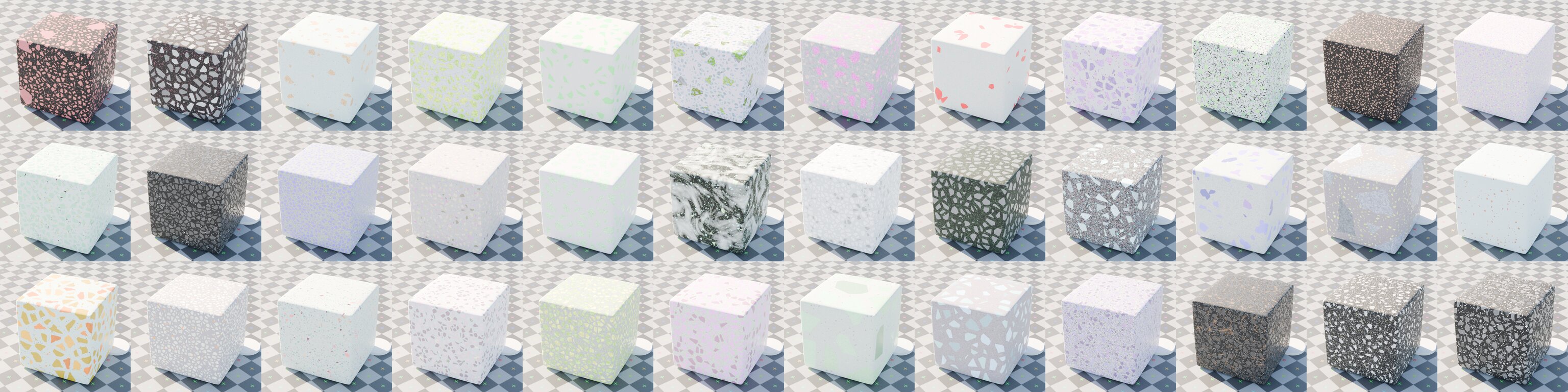}
    \vspace{-5mm}
    \caption{Terrazzo (\texttt{terrazzo\_distribution})}
    \vspace{-4mm}
\end{figure*}
\begin{figure*}[h!]
    \centering
    \vspace{-2mm}
    \includegraphics[width=\textwidth]{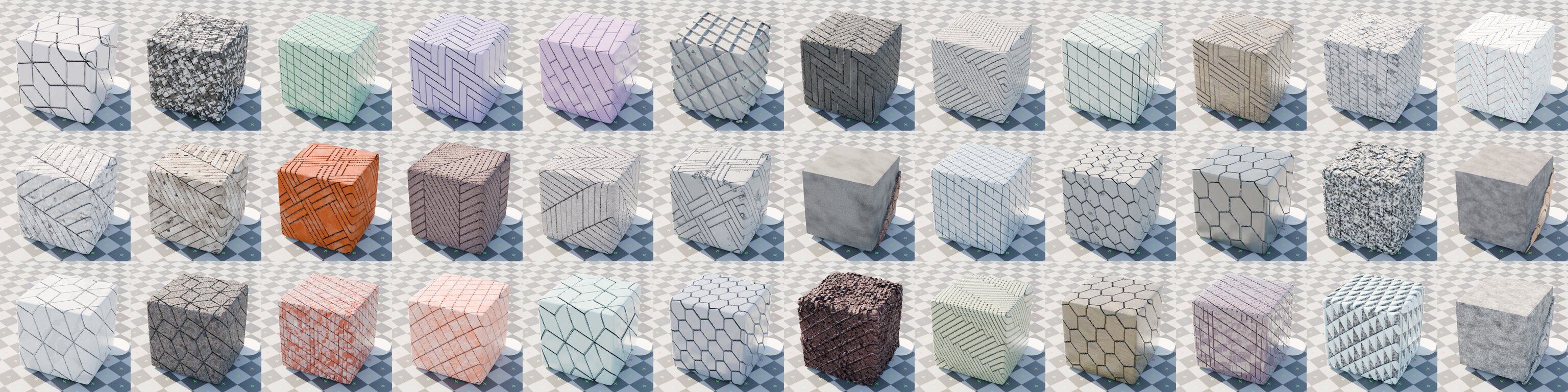}
    \vspace{-5mm}
    \caption{Tile (\texttt{tile\_distribution})}
    \vspace{-4mm}
\end{figure*}
\begin{figure*}[h!]
    \centering
    \vspace{-2mm}
    \includegraphics[width=\textwidth]{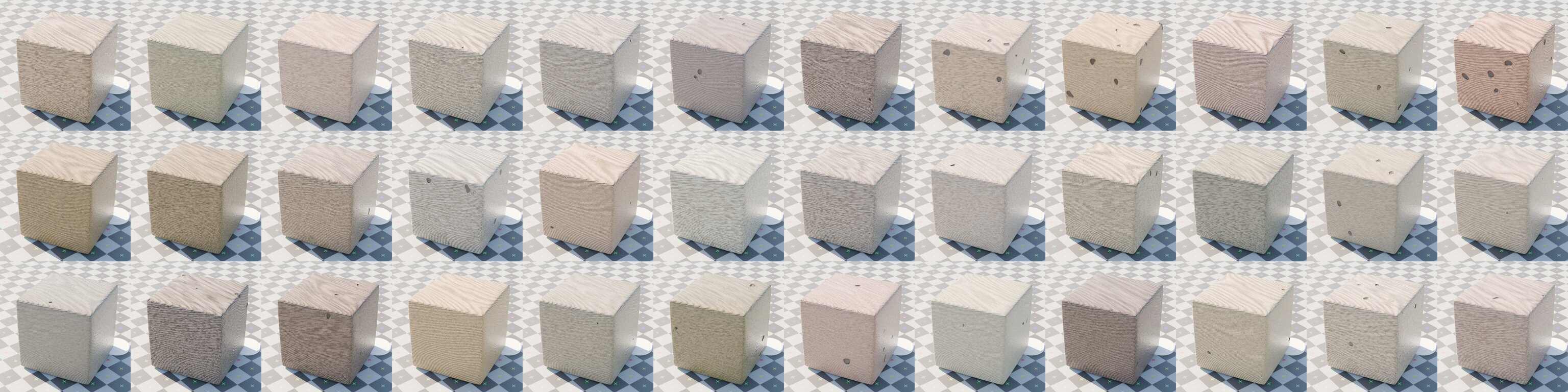}
    \vspace{-5mm}
    \caption{Wood Grain (\texttt{wood\_grain\_distribution})}
    \vspace{-4mm}
\end{figure*}

\subsection{Material Implementation Examples}
\label{sec:supp_impl}

\begin{figure}[ht]
    \caption{
        Infinigen vs ProcFunc representations of a ``Bone" material shader used by BlenderGym \cite{blendergym} from Infinigen \cite{infinigen2023infinite}. ProcFunc's is significantly shorter (73 vs 127 LOC), and uses function keyword arguments, rather than dictionary keys and Enum entries, to indicate inputs and outputs.
    }
    \centering
    \begin{subfigure}[t]{0.48\textwidth}
        \caption*{\textbf{Infinigen}}
\begin{lstlisting}[style=pythonstyle, basicstyle=\ttfamily\tiny]
def shader_bone(nw: NodeWrangler):
    # Code generated using version 2.4.3 of the node_transpiler

    texture_coordinate = nw.new_node(Nodes.TextureCoord)

    mapping = nw.new_node(
        Nodes.Mapping,
        input_kwargs={
            "Vector": texture_coordinate.outputs["Object"],
            "Location": (
                1.7 ,
                0.29999999999999999 ,
                0.0,
            ),
        },
    )

    noise_texture_2 = nw.new_node(
        Nodes.NoiseTexture,
        input_kwargs={
            "Vector": mapping,
            "Scale": 10.800000000000001,
            "Detail": 15.0,
            "Roughness": 0.76670000000000005,
        },
    )

    voronoi_texture_1 = nw.new_node(
        Nodes.VoronoiTexture,
        input_kwargs={"Vector": noise_texture_2.outputs["Fac"], "Scale": 10.0},
    )

    colorramp_2 = nw.new_node(
        Nodes.ColorRamp, input_kwargs={"Fac": voronoi_texture_1.outputs["Color"]}
    )
    colorramp_2.color_ramp.elements[0].position = 0.4364
    colorramp_2.color_ramp.elements[0].color = (0, 0, 0, 1.0)
    colorramp_2.color_ramp.elements[1].position = 0.58
    colorramp_2.color_ramp.elements[1].color = (1, 1, 1, 1.0)

    mapping_2 = nw.new_node(
        Nodes.Mapping, input_kwargs={"Vector": texture_coordinate.outputs["Object"]}
    )

    noise_texture = nw.new_node(
        Nodes.NoiseTexture,
        input_kwargs={
            "Vector": mapping_2,
            "Scale": 98.900000000000006,
            "Detail": 15.0,
            "Roughness": 0.76670000000000005,
        },
    )

    voronoi_texture = nw.new_node(
        Nodes.VoronoiTexture,
        input_kwargs={
            "Vector": noise_texture.outputs["Fac"],
            "Scale": 10.0,
        },
    )

    colorramp = nw.new_node(
        Nodes.ColorRamp, input_kwargs={"Fac": voronoi_texture.outputs["Color"]}
    )
    colorramp.color_ramp.elements[0].position = 0.3089
    colorramp.color_ramp.elements[0].color = (0.0, 0.0, 0.0, 1.0)
    colorramp.color_ramp.elements[1].position = 0.673
    colorramp.color_ramp.elements[1].color = (1.0, 1.0, 1.0, 1.0)

    multiply = nw.new_node(
        Nodes.VectorMath,
        input_kwargs={0: colorramp_2.outputs["Color"], 1: colorramp.outputs["Color"]},
        attrs={"operation": "MULTIPLY"},
    )

    mapping_1 = nw.new_node(
        Nodes.Mapping,
        input_kwargs={
            "Vector": texture_coordinate.outputs["UV"],
            "Scale": (1.0, 1.0, 0.0),
        },
    )

    noise_texture_1 = nw.new_node(
        Nodes.NoiseTexture,
        input_kwargs={
            "Vector": mapping_1,
            "Scale": 6.4000000000000004,
        },
    )

    colorramp_1 = nw.new_node(
        Nodes.ColorRamp, input_kwargs={"Fac": noise_texture_1.outputs["Fac"]}
    )
    colorramp_1.color_ramp.elements[0].position = 0.3682
    colorramp_1.color_ramp.elements[0].color = (
        0.38129999999999997,
        0.2384,
        0.1183,
        1.0,
    )
    colorramp_1.color_ramp.elements[1].position = 0.7591
    colorramp_1.color_ramp.elements[1].color = (
        0.49690000000000001,
        0.50290000000000001,
        0.46779999999999999,
        1.0,
    )

    mix = nw.new_node(
        Nodes.MixRGB,
        input_kwargs={
            "Fac": multiply.outputs["Vector"],
            "Color1": (0.19120000000000001, 0.045199999999999997, 0.0103, 1.0),
            "Color2": colorramp_1.outputs["Color"],
        },
    )

    principled_bsdf = nw.new_node(
        Nodes.PrincipledBSDF,
        input_kwargs={"Base Color": mix, "Roughness": 0.44090000000000001},
    )
\end{lstlisting}
    \end{subfigure}
    \hfill
    \begin{subfigure}[t]{0.48\textwidth}
        \caption*{\textbf{ProcFunc}}
\begin{lstlisting}[style=pythonstyle, basicstyle=\ttfamily\tiny]
def shader_bone():
    # Code generated by procfunc

    coord = pf.nodes.shader.coord()

    mapping = pf.nodes.shader.mapping(
        vector=coord.object, location=(1.6554, 0.3241, -0.007)
    )

    noise = pf.nodes.shader.noise(
        vector=mapping, scale=12.100835, detail=15.0, roughness=0.7667
    )

    voronoi = pf.nodes.shader.voronoi(
        vector=noise.fac.astype(dtype=pf.Vector), scale=10.0
    )

    color_ramp = pf.nodes.shader.color_ramp(
        fac=voronoi.color.astype(dtype=float),
        points=[(0.4, (0.0, 0.0, 0.0, 1.0)), (0.573, (1.0, 1.0, 1.0, 1.0))],
        interpolation="LINEAR",
    )

    mapping_1 = pf.nodes.shader.mapping(coord.object)

    noise_1 = pf.nodes.shader.noise(
        vector=mapping_1, scale=102.481506, detail=15.0, roughness=0.7667
    )

    voronoi_1 = pf.nodes.shader.voronoi(
        vector=noise_1.fac.astype(dtype=pf.Vector), scale=10.044822
    )

    color_ramp_1 = pf.nodes.shader.color_ramp(
        fac=voronoi_1.color.astype(dtype=float),
        points=[(0.334, (0.0, 0.0, 0.0, 1.0)), (0.717, (1.0, 1.0, 1.0, 1.0))],
        interpolation="LINEAR",
    )

    mix_rgb_factor = color_ramp.color.astype(
        dtype=pf.Vector
    ) * color_ramp_1.color.astype(dtype=pf.Vector)

    mapping_2 = pf.nodes.shader.mapping(vector=coord.uv, scale=(1.0, 1.0, 0.0))

    noise_2 = pf.nodes.shader.noise(vector=mapping_2, scale=7.187254)

    color_ramp_2 = pf.nodes.shader.color_ramp(
        fac=noise_2.fac,
        points=[
            (0.413, (0.381, 0.238, 0.118, 1.0)),
            (0.742, (0.497, 0.503, 0.468, 1.0)),
        ],
        interpolation="LINEAR",
    )

    mix_rgb = pf.nodes.shader.mix_rgb(
        factor=mix_rgb_factor.astype(dtype=float),
        a=pf.Color((0.191, 0.045, 0.01)),
        b=color_ramp_2.color,
    )

    principled = pf.nodes.shader.principled_bsdf(
        base_color=mix_rgb,
        roughness=0.4409,
        ior=1.45,
        subsurface_method="RANDOM_WALK_SKIN",
        subsurface_ior=1.4,
        subsurface_anisotropy=0.0,
        distribution="GGX",
        emission_color=pf.Color((0.0, 0.0, 0.0)),
        emission_strength=1.0,
    )
\end{lstlisting}
    \end{subfigure}
    \label{fig:supp_impl_example}
\end{figure}

\clearpage

\section{Extended Method}
\label{sec:supp_method}

\subsection{Tracer Implementation}
\label{sec:supp_tracer}

A tracer is a system which executes a program, but upon reaching certain functions, will simply record that the function would have been executed, rather than perform the actual computation. The return value of a skipped function is instead replaced with a ``Proxy" value, which records its origin point and can be passed to subsequent functions as usual. The Proxy object allows as many operations as possible, such as arithmetic and property access, so that programs which applied these operations to the non-proxy return value will still execute them correctly, albeit simply recording the operations rather than executing them. Once a Proxy has propagated all the way through the program, encountering many skipped functions along the way, it has recorded a compute graph where each yet-to-be-executed function is a graph node, and any proxy inputs and outputs are graph edges.

ProcFunc uses this framework for procedural code by decorating/marking all of its primitives and modules as functions which may need to be captured into the compute graph instead of executed. We also decorate all functions in NumPy while tracing, similarly to JAX \cite{jax2018github}. This means that a procedural generator can be analyzed without executing any graphics operations, so tracing can complete usually in less than a second. 

ProcFunc decorates all primitives, modules, random values and distribution functions, however it represents these as distinct levels of granularity of computation. Every dataset distribution is composed of random values and modules and primitives, and every module is composed of primitives. Ultimately then, a given dataset \texttt{\_distribution} function can be represented as either a single entity, or a composition of many primitives, and any one of these granularities may be useful as compute graphs.

ProcFunc's tracer allows access to any granularity via a \texttt{--trace\_level} command-line or Python argument. Its effects are demonstrated via code examples. Fig.~\ref{fig:bricks_grammar} shows the coarsest level, which can be used to generate useful starter scripts. Fig.~\ref{fig:bricks_control} shows one step more granular, which has combined the random choices and generator calls of many distributions into a single large branching graph, including random control flow. Fig.~\ref{fig:bricks_params} shows a more granular graph, which has removed all control flow to show only the path of execution that would actually be taken by a single run of the generator. Finally, Fig.~\ref{fig:bricks_specific} is the most granular, as it executes every random function instead of capturing it in the graph, and therefore has concrete values for every random choice, which provides a concrete and editable representation of a specific run of the procedural generator. These four graphs all represent the same computation, but at different levels of detail, and with different degrees of randomness either executed or left unexecuted.

\begin{figure}[ht]
    \centering
    \caption{
        \textbf{Grammar trace.} ProcFunc's Tracer can auto-generate a compute-graph which lists simply the highest level functions invoked in a computation. Here, the \texttt{bricks distribution}, \texttt{material cube} and \texttt{render cycles} functions directly correspond to the user's input command. This is the simplest case of tracing, as it does not inspect the internals of any function, but is still useful to create starting scripts which can be customized to produce randomized datasets.
    }
\begin{lstlisting}[style=pythonstyle]
# uv run procgen bricks_distribution material_cube render_cycles --codegen print --trace_level GRAMMAR

def execute_generators():
    # Code generated by procfunc v0.18.0-ifg
    rng = np.random.default_rng()

    rngs = rng.spawn(3)

    vector_vector = pf.nodes.shader.geometry()

    mapping = pf.nodes.shader.mapping(
        vector=vector_vector.position,
        rotation=(-0.7853981633974483, 0, 0),
    )

    bricks_distribution = bricks.bricks_distribution(rng=rngs[0], vector=mapping)

    material_cube = asset_demo.material_cube(rng=rngs[1], material=bricks_distribution)

    render_result = render_cycles.render_cycles(
        objects=material_cube.all_objects,
        camera=material_cube.cameras[0],
        output_folder=Path("outputs/myresult"),
        render_passes=[
            RenderPass(
                type=ExportType.IMAGE,
                path=Path("%c/%f.png"),
                data_type=np.dtype("uint8"),
            )
        ],
        frame_start=0,
        frame_end=0,
        resolution=(512, 512),
        min_samples=8,
        film_exposure=2.0,
    )
    return render_result
\end{lstlisting}
    \label{fig:bricks_grammar}
\end{figure}

\begin{figure}[ht]
    \centering
    \caption{
        \textbf{Control trace.} ProcFunc's Tracer can auto-generate a compute-graph which contains all control paths, even if they would not have been executed for the current random seed. This example includes parameters and function calls for \texttt{granite}, even though a single invocation of ``bricks'' only uses one. The true path that would be taken is instead recorded via \texttt{chosen idx}. This graph is useful for computing metrics without having to execute every possible path, or for transforming the distribution (e.g. replacing every gaussian with uniform, removing certain materials, or other effects). Ellipses (...) are code omitted for brevity.
    }
\begin{lstlisting}[style=pythonstyle]
# uv run procgen bricks_distribution material_cube render_cycles --codegen print --trace_level RANDOM_CONTROL

def execute_generators():
    # Code generated by procfunc v0.18.0-ifg
    
    ...

    brick_x = pf.random.clip_gaussian(rng=rngs_2[0], mean=0.3, std=0.1, low=0.1, high=0.5)
    brick_y = pf.random.clip_gaussian(rng=rngs_2[0], mean=0.4, std=0.1, low=0.2, high=1.0)
    brick_cutter_brick_width_vector = pf.nodes.func.combine_xyz(x=brick_x, y=brick_x * brick_y)
    brick_cutter_brick_width = brick_cutter_brick_width_vector
    brick_cutter_brick_scaling = pf.random.uniform(rng=rngs_2[2], low=1, high=2)
    brick_cutter_irregularity_choice_options_1 = pf.random.uniform(rng=rngs_2[2], low=0.3, high=0.45)
    brick_cutter_grout_bevel = pf.random.uniform(rng=rngs_2[2], low=0.4, high=1.0)
    ...
    brick_cutter_result = brick_cutter.brick_cutter(
        vector=mapping,
        brick_width=brick_cutter_brick_width.x,
        brick_height=brick_cutter_brick_width.y,
        brick_scaling=brick_cutter_brick_scaling,
        brick_bevel_concavity=brick_cutter_brick_bevel_concavity,
        ...
    )

    granite_result = granite.granite(
        ... random params ...
    )
    concrete_result = concrete.concrete(
        ... random params ...
    )

    surface = pf.control.choice(
        choice_options=[
            (brick_concrete_result.surface, 1.0),
            (granite_result.surface, 0.5),
            (concrete_result.surface, 1.0),
        ],
        chosen_idx=2,
    )
    ...
    return render_cycles.render_cycles(...)
\end{lstlisting}
    \label{fig:bricks_control}
\end{figure}

\begin{figure}[ht]
    \centering
    \caption{
        \textbf{Params trace.} ProcFunc's Tracer can auto-generate a compute-graph which contains only the \textit{active} parameters for a particular execution of a randomized generator. Unlike Fig.~\ref{fig:bricks_control}, this code does not contain \texttt{granite}, since granite is not used in this particular execution of the bricks. The params trace of a material or object can be used to know the continuous distribution of that object, or to optimize the parameters of the object to meet a target while remaining within the support of the distribution. Ellipses (...) are code omitted for brevity.
    }
\begin{lstlisting}[style=pythonstyle]
# uv run procgen bricks_distribution material_cube render_cycles --codegen print --trace_level RANDOM_PARAMS

def execute_generators():
    # Code generated by procfunc v0.18.0-ifg
    
    ...

    brick_x = pf.random.clip_gaussian(rng=rngs_2[0], mean=0.3, std=0.1, low=0.1, high=0.5)
    brick_y = pf.random.clip_gaussian(rng=rngs_2[0], mean=0.4, std=0.1, low=0.2, high=1.0)
    brick_cutter_brick_width_vector = pf.nodes.func.combine_xyz(x=brick_x, y=brick_x * brick_y)
    brick_cutter_brick_width = brick_cutter_brick_width_vector
    brick_cutter_brick_scaling = pf.random.uniform(rng=rngs_2[2], low=1, high=2)
    brick_cutter_irregularity_choice_options_1 = pf.random.uniform(rng=rngs_2[2], low=0.3, high=0.45)
    brick_cutter_grout_bevel = pf.random.uniform(rng=rngs_2[2], low=0.4, high=1.0)
    ...
    brick_cutter_result = brick_cutter.brick_cutter(
        vector=mapping,
        brick_width=brick_cutter_brick_width.x,
        brick_height=brick_cutter_brick_width.y,
        brick_scaling=brick_cutter_brick_scaling,
        brick_bevel_concavity=brick_cutter_brick_bevel_concavity,
        ...
    )
    rngs_3 = rngs_1[2].spawn(2)
   
    concrete_base_color_hue = pf.random.uniform(rng=rngs_1[1], low=np.array([0., 0.]), high=np.array([1. , 0.1]))
    concrete_base_color_value = pf.random.clip_gaussian(rng=rngs_1[1], mean=0.5, std=0.2, low=0.2, high=0.95)
    concrete_base_color = pf.color.hsv_color(
        hue=concrete_base_color_hue[0],
        saturation=concrete_base_color_hue[1],
        value=concrete_base_color_value,
    )
    concrete_roughness = pf.random.uniform(rng=rngs_1[1], low=0.5, high=1.0)
    ...
    surface = concrete.concrete(
        vector=brick_cutter_result.vector,
        base_color=concrete_base_color,
        roughness=concrete_roughness,
        ...
    )
    ...
    return render_cycles.render_cycles(...)
\end{lstlisting}
    \label{fig:bricks_params}
\end{figure}

\begin{figure}[ht]
    \centering
    \caption{
        \textbf{Specific trace.} ProcFunc's Tracer can auto-generate a compute-graph or Python file which expresses the specific numerical parameters which correspond to a particular execution of a randomized procedural generator. Ellipses (...) are code omitted for brevity. This means that randomized procedural generators can automatically emit code for non-randomized procedural generators for each sample. This is useful for editing specific examples, either via the generated Python code or the underlying compute graph it was generated from. It is also plausibly useful for LLM training, by creating many diverse image-text pairs, even with differing code structure due to random control paths.
    }
\begin{lstlisting}[style=pythonstyle]
# uv run procgen bricks_distribution material_cube render_cycles --codegen print --trace_level GENERATORS

def execute_generators():
    # Code generated by procfunc v0.18.0-ifg

    ...

    brick_cutter_brick_width = pf.nodes.func.combine_xyz(
        x=0.14953272082744012, y=0.05129953938152119
    )
    brick_cutter_result = brick_cutter.brick_cutter(
        vector=mapping,
        brick_width=brick_cutter_brick_width.x,
        brick_height=brick_cutter_brick_width.y,
        brick_scaling=1.9920081006344654,
        brick_bevel_concavity=0.11666851003155587,\
        ...
    )
    ...
    concrete_base_color = pf.color.hsv_color(
        hue=0.6258491463323542,
        saturation=0.0245507562345461,
        value=0.4491897512413083,
    )

    concrete_micro_grain = pf.control.choice(
        choice_rng=rngs_1[1],
        choice_options=[(0.5572780986583625, 1.0), (0.0, 1.0)],
        chosen_idx=0,
    )
    surface = concrete.concrete(
        vector=brick_cutter_result.vector,
        base_color=concrete_base_color,
        roughness=0.8040807062831756,
        ...
    )
    ...
    return render_cycles.render_cycles(...)
\end{lstlisting}
    \label{fig:bricks_specific}
\end{figure}

\clearpage

\subsection{Base Generator}
\label{sec:supp_basegen}

\subsubsection{Efficient Wall Geometry}
\label{sec:supp_walls}
The walls, floor and ceiling of an indoor room present a very large surface area which must be densely subdivided to achieve high quality mesh displacement and accurate geometric ground truth. Whereas Infinigen Indoors \cite{infinigen2024indoors} uses either no dense geometry or dense geometry from volumetric marching cubes, our system can create dense mesh geometry only through mesh surface subdivision. Infinigen-Indoors uses mesh booleaning to create openings for doors and windows, which leads to complex, non-convex, non-quadrilateral polygons, which are not amenable to mesh subdivision. To avoid this, we treat room planes as parametric UV surfaces, and create holes for windows or other protrusions without any booleaning. To add a set of windows at a set of locations, we calculate the U and V parameters for the corners of the windows, and simply decline to generate any polygons within that U and V range. This results in mesh geometry which has even quadrilateral topology while still leaving holes for the windows. We then add a Blender subdivision operation, but leave it in an unapplied state, where it is then efficiently executed and displaced by the shader graph upon render startup.

\subsubsection{Object Generators}
\label{sec:supp_objects}

We derive the current library of object-level procedural generators via refactoring generators already present in Infinigen-Indoors \cite{infinigen2024indoors}. This is \textit{unlike} our material assets and scene-arrangement tools, which are wholly original. So far, we have refactored \textit{Bookcases, Cabinets, Ceiling lights, Desks, Drawers, Flowers, Lamps, Sofas, Tables, Triangle Shelves, Vases, Wall Art, Mirrors and Windows} to use ProcFunc Primitives and Module Interfaces. This involves significant engineering and improvement to the generators - we partially automate the process for the non-randomized components by using the transpiler, but have manually re-implemented and adjusted the random parameter sampling code to use the new traceable ProcFunc standard. We also create weighted lists of materials to be used for various purposes within these objects (e.g. a tabletop may use metal, wood, planks, marble). We also create the needed UV unwrappings to apply our materials.

Besides refactored procedural generators, we also demonstrate that ProcFunc can import arbitrary external assets and use them efficiently in scene arrangement. Particularly we do this for small object assets, which we pre-generate with Infinigen Indoors \cite{infinigen2024indoors} from a variety of existing asset classes: \textit{Books, Bottles, Bowls, cans, Chopsticks, Cups, Food Bags, Boxes, Forks, Jars, Knives, Pans, Plates, Pots, Spoons, Towels, Vases, Wine glasses, Large Plants, Small Plants}.

To do this, we generated and exported a library of these assets in obj format. We then filtered out the objects that were over 100MB and only imported ones that were under the limit. Our final library consisted of 279 pre-generated assets. 

Infinigen Indoors \cite{infinigen2024indoors} does not come with functionality to import and place external assets. Therefore, we implemented asset importing and instancing utilities. The object importer takes a folder of pre-generated assets and imports them into the scene, supporting \verb|dae, abc, usd, obj, ply, stl, fbx, glb, gltf| and \verb|blend| formats. If the object consists of a tree, the importer collapses it into a single mesh. The user defines an external asset distribution using the following API, which samples a random pre-generated asset from the folder: \verb|object_distribution = distribution_from_asset_glob(folder)|

The instancing utility takes a parent object and a child object, and instances the child object on top of the parent object using a Poisson disk distribution. This allows us to distribute a large number of small objects while keeping the memory usage down. It finally randomizes the pose of each instance and checks for collisions. 

\subsubsection{Collision}
\label{sec:supp_collision}

We implement a collision API that allows for efficient collision testing while being simple from a user perspective. A minimal example of the API is as follows: 
\begin{verbatim}
col = collision_set(objs, existing_set)   
collides = intersection_test(col, query_obj):
hits_xyz, index_ray, index_tri = raycast(col, ray_origins, ray_dirs)
\end{verbatim}
We use the Trimesh and the FCL libraries to manage the collisions. Every mesh builds an FCL BVH once and every instance is mapped to its source mesh. Thus, we do not store separate meshes for instances. This is crucial in terms of efficient runtime and memory since most small objects are instanced. 

\subsubsection{Ground Truth}
\label{sec:supp_gt}

Infinigen's \cite{infinigen2023infinite} ground-truth system is tightly coupled with its scene generation code. This means that the user cannot take an arbitrary Blender scene and export the ground-truth from it. Furthermore, Infinigen does not allow the user to customize the ground-truth exports in terms of file type, path, and which types of ground-truth to choose. 

Motivated by this, we implement our own ground-truth API, which decouples ground-truth exports from scene generation and allows the user to customize what type of ground-truth they want. Here is an example usage of our API:

\begin{lstlisting}[language=Python, breaklines]
    objects, camera = configure_scene()

    render_passes = [
        RenderPass(ExportType.DEPTH, Path("depth.npy"), np.dtype(np.float32)),
        RenderPass(ExportType.SURFACE_NORMAL, Path("normal.npy"), np.dtype(np.float32)),
        RenderPass(ExportType.OPTICAL_FLOW, Path("flow.npy"), np.dtype(np.float32)),
    ]

    if method == "cycles":
        results = render_cycles_ground_truth(
            objects=objects,
            camera=camera,
            output_folder=tmp_path,
            render_passes=render_passes,
            frame_start=1,
            frame_end=10,
            resolution=(256, 256),
        )
    elif method == "eevee":
        results = render_eevee_ground_truth(
            objects=objects,
            camera=camera,
            output_folder=tmp_path,
            render_passes=render_passes,
            frame_start=1,
            frame_end=10,
            resolution=(256, 256),
        )
\end{lstlisting}

Note that our API can be called in any Blender scene, not just ones generated from our system. The filetypes we support include \verb|.png, .exr, .npy, .npz|. 

Infinigen uses the Cycles rendering engine to obtain most of its ground-truth types and to render the RGB images. Since Cycles is a physically-based rendering engine, this process is slow. Thus, we implement an additional API to render and export ground-truth using the EEVEE engine, which is a rasterization-based rendering engine. EEVEE is much faster than Cycles, especially for scenes with simple materials. The user can choose which engine to use for their renders and ground-truth. 

\subsubsection{RRT* Camera Trajectory}
\label{sec:supp_rrt}
Infinigen \cite{infinigen2023infinite} uses a random-walk-based camera trajectory generation algorithm. The issue is that this algorithm lacks diversity and does not generate interesting trajectories. Hence, we implement an RRT* \cite{LaValle1998RapidlyexploringRT,karaman_sampling-based_2011}-based camera trajectory generation algorithm. Our algorithm takes a start and a goal point as input. At each iteration, it samples a random point, finds the nearest neighbor in the existing tree, and steers towards the random point from the neighbor. It chooses a distance-minimizing parent from the neighborhood of the new step, and adds the new point to the tree. It also uses raycasting to test if new nodes are valid points in the sense that there are no obstacles in the way. RRT* allows us to generate much more diverse and dynamic trajectories.

\section{Extended Experiments}
\label{sec:supp_exp}

\begin{figure}[p]
\centering
\begin{lstlisting}[style=prompt]
# Prompt for run_llm_inference.py with procfunc backend (paramedit)
# Images: first=current render, second=goal render
#
The following code uses the `procfunc` API to produce a material in Blender:
```python
{code}
```

Your job is to edit the PARAMETERS of the code to make the render more like the goal material. Only edit the parameters, not the structure or args.

Answer the following questions:
1) What is the SINGLE most visually obvious difference between the two materials shown below?
2) Look at the code. Which fields/variables which are set to numerical values are most likely responsible for the obvious visual difference in your answer to question 1?
3) Copy the code above (COPY ALL OF IT) and replace the assignments of such fields/variables accordingly!

- MAKE SURE YOUR CODE IS RUNNABLE. You do not need to add your material to an object, it will be auto-rendered so long as you assign it to `shader` at the end of the file
- Only use functions inside `pf.nodes`, not any other way of accessing blender. These functions have a similar interface to standard blender nodes.
- DO NOT BE BRIEF IN YOUR CODE. DO NOT ABBREVIATE YOUR CODE WITH "..." -- TYPE OUT EVERYTHING.
- Return your code in a ```python code block.
\end{lstlisting}
\caption{Prompt template for the \texttt{procfunc} backend in the \textbf{parameter-edit} setting. The \texttt{\{code\}} placeholder is replaced with the current code at inference time. Two images are provided: the current render and the goal render. This prompt is closely based on BlenderGym's, but has been modified for use with our evaluation setup.}
\label{fig:prompt-pf-paramedit}
\end{figure}

\begin{figure}[p]
\centering
\begin{lstlisting}[style=prompt]
# Prompt for run_llm_inference.py with bgym/infinigen backend (paramedit)
# Images: first=current render, second=goal render
#
The following Blender code was used to produce a material:
```python
{code}
```

Your job is to edit the PARAMETERS of the code to make the render more like the goal material. Only edit the parameters, not the structure or args.

Answer the following questions:
1) What is the SINGLE most visually obvious difference between the two materials shown below?
2) Look at the code. Which fields/variables which are set to numerical values are most likely responsible for the obvious visual difference in your answer to question 1?
3) Copy the code above (COPY ALL OF IT) and replace the assignments of such fields/variables accordingly!

- MAKE SURE YOUR CODE IS RUNNABLE. MAKE SURE TO ASSIGN THE FINAL MATERIAL TO `material_obj` (through `apply(material_obj)`) AS THE LAST LINE OF YOUR CODE.
- DO NOT BE BRIEF IN YOUR CODE. DO NOT ABBREVIATE YOUR CODE WITH "..." -- TYPE OUT EVERYTHING.
- Return your code in a ```python code block.
\end{lstlisting}
\caption{Prompt template for the \texttt{infinigen} backend in the \textbf{parameter-edit} setting. The \texttt{\{code\}} placeholder is replaced with the current code at inference time. Two images are provided: the current render and the goal render. This prompt is closely based on BlenderGym's, but has been modified for use with our evaluation setup.}
\label{fig:prompt-ifg-paramedit}
\end{figure}

\begin{figure}[p]
\centering
\begin{lstlisting}[style=prompt]
# Prompt for run_llm_inference.py with procfunc backend (fromscratch)
# Images: only goal render is provided
#
The following code uses the `procfunc` API to produce a material in Blender:
```python
{code}
```

The image shows the desired goal material. Please edit the code above to produce a material that matches this goal.

Pay special attention to the base color of the materials, as well as any patterns, textures, or details.

- MAKE SURE YOUR CODE IS RUNNABLE. You do not need to add your material to an object, it will be auto-rendered so long as you assign it to `shader` at the end of the file
- Only use functions inside `pf.nodes`, not any other way of accessing blender. These functions have a similar interface to standard blender nodes.
- DO NOT BE BRIEF IN YOUR CODE. DO NOT ABBREVIATE YOUR CODE WITH "..." -- TYPE OUT EVERYTHING.
- Return your code in a ```python code block.
\end{lstlisting}
\caption{Prompt template for the \texttt{procfunc} backend in the \textbf{from-scratch} setting. The \texttt{\{code\}} placeholder is replaced with starter code at inference time. A single image (the goal render) is provided. This prompt is closely based on BlenderGym's, but has been modified for use with our evaluation setup.}
\label{fig:prompt-pf-fromscratch}
\end{figure}

\begin{figure}[p]
\centering
\begin{lstlisting}[style=prompt]
# Prompt for run_llm_inference.py with bgym/infinigen backend (fromscratch)
# Images: only goal render is provided
#
The following Blender code was used to produce a material:
```python
{code}
```
The final material is assigned to the object `material_obj`, a sphere.

The image shows the desired goal material. Please edit the code above to produce a material that matches this goal.

Pay special attention to the base color of the materials, as well as any patterns, textures, or details.

MAKE SURE YOUR CODE IS RUNNABLE. MAKE SURE TO ASSIGN THE FINAL MATERIAL TO `material_obj` (through `apply(material_obj)`) AS THE LAST LINE OF YOUR CODE.
DO NOT BE BRIEF IN YOUR CODE. DO NOT ABBREVIATE YOUR CODE WITH "..." -- TYPE OUT EVERYTHING.
Return your code in a ```python code block.
\end{lstlisting}
\caption{Prompt template for the \texttt{infinigen} backend in the \textbf{from-scratch} setting. The \texttt{\{code\}} placeholder is replaced with starter code at inference time. A single image (the goal render) is provided. This prompt is closely based on BlenderGym's, but has been modified for use with our evaluation setup.}
\label{fig:prompt-ifg-fromscratch}
\end{figure}

\subsection{LLM Editing}
\label{sec:supp_llm}
We provide exact prompts used for our parameter editing (Fig.~\ref{fig:prompt-ifg-paramedit}, \ref{fig:prompt-pf-paramedit}) and from-scratch procedural generator creation settings (Fig.~\ref{fig:prompt-ifg-fromscratch},\ref{fig:prompt-pf-fromscratch}). These are based on BlenderGym\cite{blendergym} with as little adaptation as possible to allow use with ProcFunc and with our new setting. 

\clearpage

\begin{figure}[th]
    \centering
    \caption{
       \textbf{Maximum Diversity Composition} - We provide a random sampling function which combines together all possible arrangements of ProcFunc's material components, rather than just the more ordinary combinations in Fig.\ref{fig:qual_realistic_firstmat} to \ref{fig:qual_realistic_lastmat}. This enables a very large number of combinations, including the particular combination of any two materials with any shape and including or not including every type of wear and tear or added layer. 
    }
    \includegraphics[width=\textwidth]{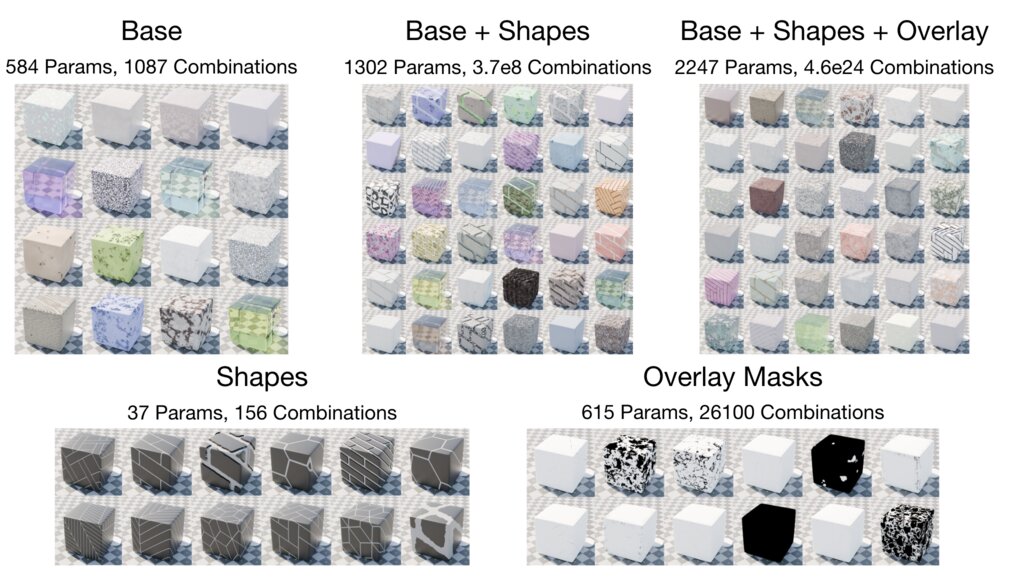}
    \label{fig:diversity}
\end{figure}

\subsection{Procedural Generator Diversity}
\label{sec:supp_diversity}

In Figure~\ref{fig:diversity} we display samples from a material generator constructed to be extremely random and compositional. Unlike the previous material generators tuned for realism, this generator aims to demonstrate the composition modes of base materials, shapes, and overlay masks. We use the tracer to generate a compute graph of the generator down to the level of individual random choices. From this graph, we calculate the total number of continuous parameters and the number of possible combinations from the discrete choices. For example, when combining base materials with shapes, we can have bricks with an arbitrary ``brick'' material and an arbitrary ``grout'' material. This creates more than 1 million combinations within bricks alone. Overall, we have $4.6 \times 10^{24}$ possible combinations after composing base materials, shapes, and overlay masks. 

\subsection{Procedural Generator Efficiency}
\label{sec:supp_efficiency}
The efficiency experiments measure performance and complexity across 10 different settings: 
\begin{itemize}
    \item \textbf{Infinigen Indoors \cite{infinigen2024indoors} without objects and lowpoly:} We only generate rooms and disable object generation. We use the default settings of Infinigen Indoors and do not use OcMesher \cite{ocmesher2023view}. 
    \item \textbf{Our system without objects and lowpoly:} We only generate rooms and do not place objects. We disable subdivisions in room generation. 
    \item \textbf{Our system without objects, lowpoly, and rendered with EEVEE:} Same as the previous setting, except we render with the rasterization-based EEVEE rendering engine and use EEVEE ground truth instead of the raytracing-based Cycles. We realize meshes before rendering. 
    \item \textbf{Infinigen Indoors \cite{infinigen2024indoors} without objects and hipoly:} We only generate rooms and disable object generation. We use the default settings of Infinigen Indoors and use OcMesher \cite{ocmesher2023view} to get higher detail rooms.
    \item \textbf{Our system without objects and hipoly:} We only generate rooms and do not place objects. We enable subdivisions during room generation. 
    \item \textbf{Infinigen Indoors \cite{infinigen2024indoors} with objects and lowpoly:} We generate full livingrooms in infinigen and enable object generation. We use default settings and do not use OcMesher \cite{ocmesher2023view}. 
    \item \textbf{Our system with objects and lowpoly:} We generate livingrooms with objects and disable subdivisions. The subdivisions are disabled for both room generation and object generation. 
    \item \textbf{Our system with objects and lowpoly, rendered with EEVEE:} We generate livingrooms with objects and disable subdivisions. The subdivisions are disabled for both room generation and object generation. We realize meshes before rendering and use the EEVEE rendering engine instead of Cycles. 
    \item \textbf{Infinigen Indoors full:} We generate Infinigen Indoors livingrooms with objects and use OcMesher \cite{ocmesher2023view} to subdivide the rooms. 
    \item \textbf{Our system full:} We generate livingrooms with objects and enable subdivisions. We render with the Cycles rendering engine. 
\end{itemize}

Our system can run as a single job in a Slurm pipeline. We use 4 Intel(R) Xeon(R) Gold 5320 CPU cores, 48 GB of memory, and 1 NVIDIA L40 GPU per run, i.e. a single livingroom generation. 

Infinigen Indoors \cite{infinigen2024indoors} runs in multiple jobs. We use 4 Intel(R) Xeon(R) Gold 5320 CPU cores and 24 GB of memory for the scene generation stage,  8 CPU cores, 48 GB memory, and 1 L40 GPU for rendering. There is also a backup rendering stage if initial rendering fails which uses 16 CPU cores, 96 GB of memory, and 2 L40 GPUs. The ground-truth stage uses 8 CPU cores and 48 GB of memory.

In \cref{tab:gt_perf} we report ground-truth generation times. Our full system is much faster than full Infinigen Indoors ground-truth generation. 

\begin{table}
  \caption{Ground-truth render runtimes for Infinigen Indoors and our system. GT Render $t_1+t_2$ means that the first frame takes $t_1$ seconds to generate and the remaining frames take $t_2$ seconds per frame.}
  \label{tab:gt_perf}
  \centering
  \begin{tabular}{@{}lccr@{}}
  \toprule
  \multicolumn{3}{c}{Settings} & Performance \\
  \cmidrule(lr){1-3}\cmidrule(lr){4-4}
  Method & Obj? & Hipoly? & GT Render $\downarrow$ \\
  \midrule
  InfIndoors & $\times$ & $\times$ & \textbf{7.65}$+$5.23s \\
  Ours & $\times$ & $\times$ & 10.46$+$\textbf{2.40s} \\
  Ours Eevee & $\times$ & $\times$ & 71.99$+$4.15s \\
  \hdashline
  InfIndoors & $\times$ & $\checkmark$ & 938.13$+$8.74s \\
  Ours & $\times$ & $\checkmark$ & \textbf{25.86$+$2.91s} \\
  \hdashline
  InfIndoors & $\checkmark$ & $\times$ & 196.17$+$6.48s \\
  Ours & $\checkmark$ & $\times$ & \textbf{13.12$+$2.91s} \\
  Ours Eevee & $\checkmark$ & $\times$ & 105.79$+$5.21s \\
  \hdashline
  InfIndoors & $\checkmark$ & $\checkmark$ & 356.68$+$9.19s \\
  Ours & $\checkmark$ & $\checkmark$ & \textbf{30.76$+$3.80s} \\
  \bottomrule
  \end{tabular}
\end{table}

\clearpage

\end{document}